\documentclass[10pt,twocolumn,letterpaper]{article}

\usepackage{cvpr}              

\usepackage{graphicx}
\usepackage{amsmath}
\usepackage{amssymb}
\usepackage{paralist}
\usepackage{booktabs}       
\usepackage{xcolor}         
\usepackage{multirow}
\usepackage[normalem]{ulem}
\usepackage{float}
\usepackage{makecell}
\usepackage{algorithm}
\usepackage{algpseudocode}
\usepackage{pifont}
\usepackage{bbding}

\usepackage[pagebackref,breaklinks,colorlinks]{hyperref}

\usepackage[capitalize]{cleveref}
\crefname{section}{Sec.}{Secs.}
\Crefname{section}{Section}{Sections}
\Crefname{table}{Table}{Tables}
\crefname{table}{Tab.}{Tabs.}

\begin{document}
	
	\title{SATB-VR: Training Few-Step Video Restoration Diffusion Model using SNR-Aware Trajectory Blending}
	
	\author{
        \vspace{-10mm} \\
		Haoran Bai\textsuperscript{1,*}  \quad
        Xiaoxu Chen\textsuperscript{1,*}  \quad
        Xiaoyu Liu\textsuperscript{1,2}  \quad
        Zongsheng Yue\textsuperscript{3}  \\
        Sibin Deng\textsuperscript{1,$\dagger$}  \quad
        Wangmeng Zuo\textsuperscript{2}  \quad
        Ying Chen\textsuperscript{1,$\dagger$}\\
		\textsuperscript{1}Alibaba Group  ~~~ \textsuperscript{2}Harbin Institute of Technology  ~~~ \textsuperscript{3}Xi'an Jiaotong University \\
		\url{https://github.com/chenxx89/SATB-VR}
	}
	
	\twocolumn[{
		\maketitle
		\vspace{-8mm}
		\begin{figure}[H]
			\hsize=\textwidth
			\centering
			\includegraphics[width=2.08\linewidth]{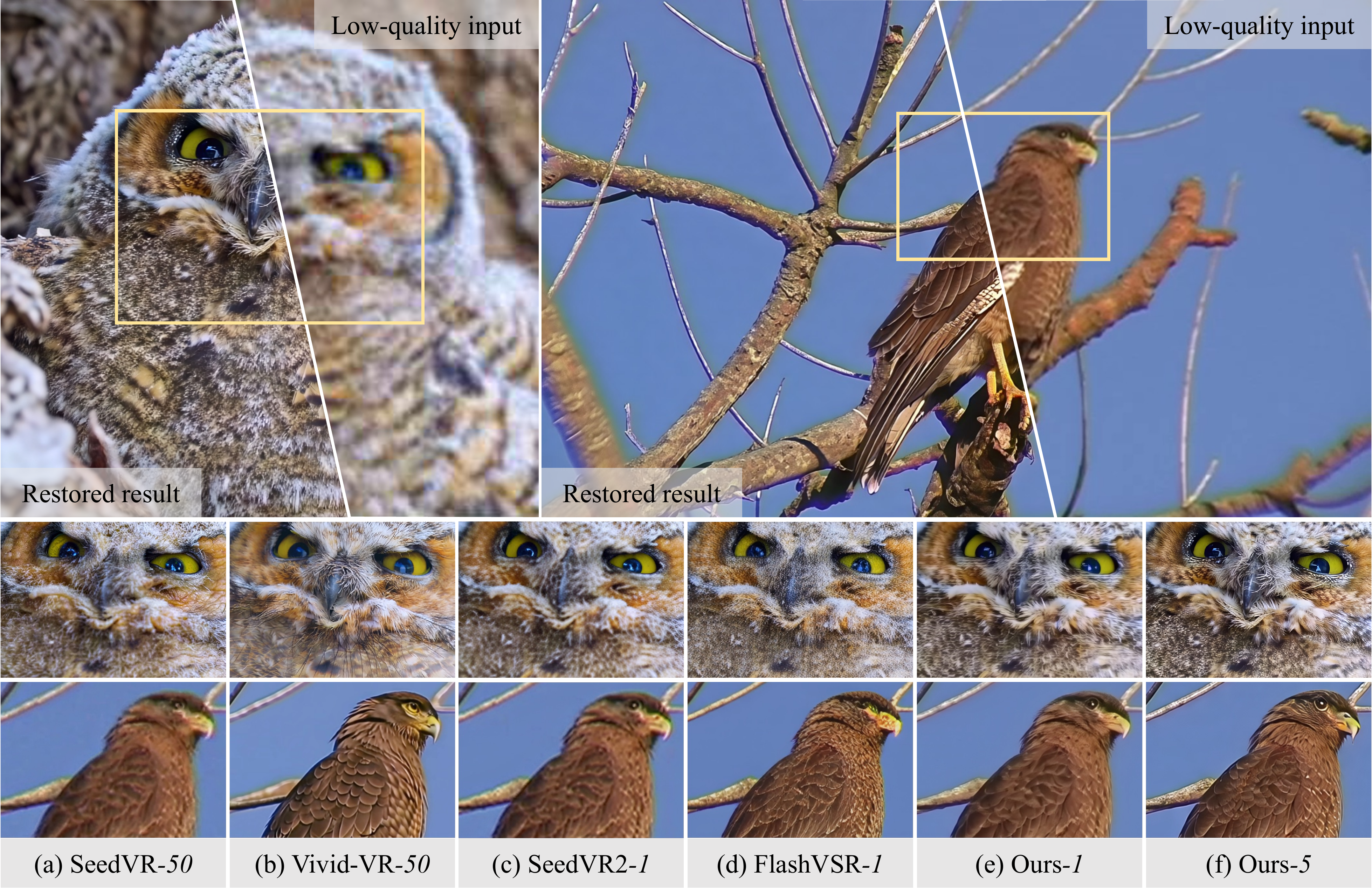}
			\vspace{2mm}
			\caption{%
				Visual comparison of video restoration. 
				Aggressive \textit{1-step} methods lack iterative correction, often causing over-smoothed results or artifacts. Recognizing that a few iterative steps are inherently required for complex detail recovery, the proposed method uses no more than \textit{5 steps} to achieve comparable or even better performance compared with \textit{50-step} approaches.
				(\textbf{Zoom-in for best view})
			}
			\label{fig:teaser}
			\vspace{3mm}
		\end{figure}
	}]

    \newcommand\blfootnote[1]{%
      \begingroup
      \renewcommand\thefootnote{}\footnote{#1}%
      \addtocounter{footnote}{-1}%
      \endgroup
    }
    \blfootnote{\textsuperscript{*} Equal contribution. \quad \textsuperscript{$\dagger$} Corresponding authors.}

	
    \vspace{-5mm}
	\begin{abstract}
		\vspace{-3mm}
		While diffusion models excel in video restoration, their reliance on extensive iterative steps limits efficiency. Conversely, aggressive single-step distillation often compromises fine texture recovery.
		To achieve an optimal balance, we present SATB-VR, a few-step paradigm that jump-starts the denoising process via an auxiliary predictor, explicitly bypassing early low signal-to-noise ratio (SNR) steps. However, naive joint training of the predictor and the denoiser inherently introduces a severe train-inference discrepancy. To resolve this, we propose the SNR-Aware Trajectory Blending (SATB) strategy.
		During the forward process, SATB constructs the noisy input by dynamically blending the predictor's output with the ground-truth trajectory based on the SNRs. This forces the denoiser to robustly compensate for initial prediction errors while smoothly converging to the clean data manifold.
		Furthermore, we introduce a Denoiser-Driven Consistency (DDC) loss, leveraging the concurrently updated denoiser as a dynamic evaluator to explicitly align internal features and boost predictor accuracy.
		Extensive experiments demonstrate that, under flexible few-step inference regimes (\eg, $\le 5$ steps), SATB-VR performs favorably against existing approaches on synthetic, real-world, and AIGC benchmarks.
	\end{abstract}
	
	
	\vspace{-5mm}
	\section{Introduction}
	\label{sec:introduction}
	\vspace{-2mm}

	\begin{figure*}[!t]
		\centering
		\includegraphics[width=1.0\linewidth]{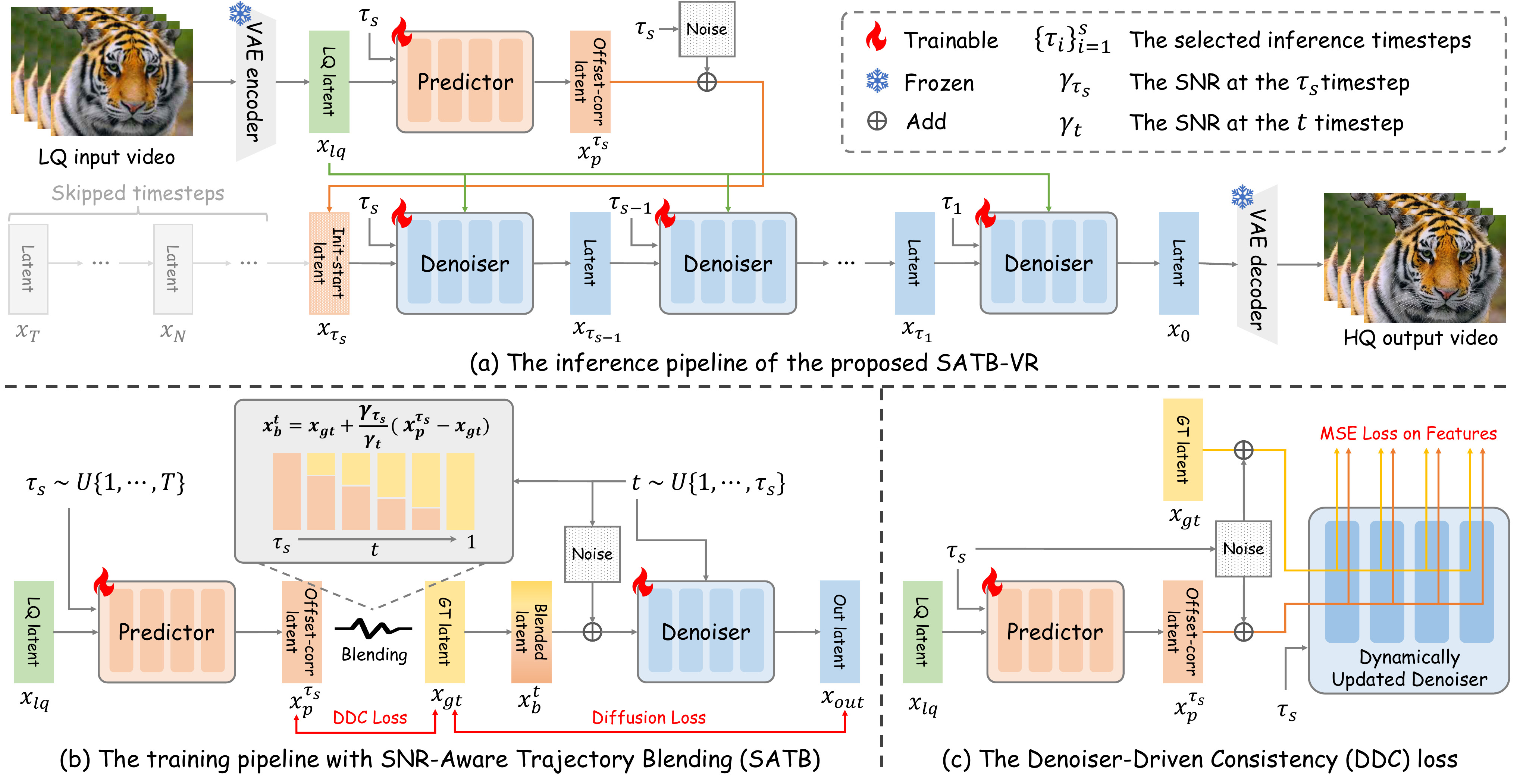}
		\caption{%
			An overview of the proposed method. (a) The inference pipeline. Given the selected timesteps $\{\tau_i\}_{i=1}^s$, the auxiliary predictor first estimates the initial state at $\tau_s$. The conditional video denoiser then iteratively performs denoising over $s$ steps for video restoration. (b) The joint training pipeline. By dynamically blending the predictor's output with the ground-truth trajectory based on the SNRs, the proposed SNR-Aware Trajectory Blending (SATB) strategy effectively eliminates the train-inference discrepancy. (c) The Denoiser-Driven Consistency (DDC) loss, utilizing the dynamically updated denoiser as a feature evaluator to explicitly constrain the predictor.
		}
		\label{fig:framework}
	\end{figure*}

	While diffusion-based methods~\cite{wang2025seedvr,xie2025star,bai2025vivid} have achieved state-of-the-art performance in video restoration, their reliance on extensive iterative denoising (\eg, $50$ steps) leads to prohibitive computational overhead. At the other extreme, aggressively distilling these models into single-step generators~\cite{wang2025seedvr2,chen2025dove,zhuang2025flashvsr,lv2026duo} collapses the sampling trajectory into a single feed-forward pass. Without progressive error correction, such single-step methods struggle to handle complex spatiotemporal degradations, often resulting in over-smoothed details and temporal flickering. Thus, the few-step paradigm provides a more balanced solution by retaining a minimal yet essential iterative process.

	Although prevailing few-step acceleration techniques, such as trajectory distillation, can significantly reduce sampling steps, they inevitably impair the synthesis of fine high-frequency details when pushed to extreme low-step regimes. Since restoration tasks primarily aim to recover high-frequency textures from degraded inputs, generating coarse structures from scratch is largely redundant. Thus, directly bypassing the early low signal-to-noise ratio (SNR) timesteps emerges as a more sound alternative to compressing the entire trajectory. Leveraging this insight, recent work~\cite{yue2025arbitrary} based on diffusion inversion employs an auxiliary predictor to estimate intermediate latents from degraded inputs. By explicitly initializing the diffusion process at high-SNR timesteps while keeping the entire diffusion backbone fixed, it achieves image restoration within $5$ steps.
	
	However, extending this paradigm to video restoration is highly non-trivial. A frozen diffusion backbone poses two critical limitations: (1) it fails to incorporate the low-quality (LQ) video as a necessary control condition, and (2) it cannot adaptively correct predictor-induced errors. These flaws become more pronounced under the complex spatiotemporal degradations, necessitating the construction of a conditional denoiser and jointly optimizing it with the auxiliary predictor. Yet, this joint training introduces a fundamental optimization dilemma. Since inference jump-starts from the predictor’s imperfect estimation, training the denoiser purely on ground-truth (GT) latents leaves it incapable of correcting these initial errors. Conversely, naively replacing the GT latent with the predictor's output during the forward training process disrupts the underlying diffusion manifold. This prevents the model from converging to the clean data distribution, causing a train-inference discrepancy and producing noticeable visual artifacts (see Fig.~\ref{fig:visual-artifact}(b)).

	To address this issue, we propose a simple yet effective training strategy, termed SNR-Aware Trajectory Blending (SATB). During joint optimization, rather than constructing the forward process solely on the GT latent, SATB injects noise into a dynamic blend of the GT latent and the predictor's output. To simulate jump-starting, we simultaneously sample a starting timestep $\tau_s$ and a current diffusion step $t \le \tau_s$. Crucially, the blending is modulated by their relative SNRs. At $t = \tau_s$, the predictor's output dominates the blend, forcing the denoiser to adaptively compensate for estimation errors. As $t$ decreases toward high-SNR timesteps, this blending weight dynamically decays, smoothly guiding the trajectory back to the clean data manifold. By anchoring the optimization objective to the GT latent, the denoiser successfully rectifies predictor-induced errors while preserving robust iterative inference capabilities.

	Furthermore, we introduce a new loss function, termed Denoiser-Driven Consistency (DDC) loss. It leverages the concurrently optimized denoiser as a dynamic feature evaluator. By explicitly aligning internal denoising features, DDC significantly enhances the predictive accuracy of the auxiliary predictor. Benefiting from these proposed strategies, we develop a generative video restoration model built upon T2V foundation model~\cite{yang2024cogvideox}, termed SATB-VR. Supporting flexible few-step inference (\eg, $\le 5$ steps), it achieves comparable or even better performance compared with existing $50$-step approaches (see Fig.~\ref{fig:teaser}). In summary, our main contributions are as follows:
	\begin{compactitem}
		\item We propose SATB, a simple yet effective training strategy that resolves the train-inference discrepancy, which enables robust joint optimization for the jump-starting video restoration paradigm.
		\item We introduce DDC, a loss function that leverages the dynamically updated denoiser as an evaluator to explicitly align internal features, thereby significantly improving the accuracy of the predictor.
		\item We present SATB-VR, with no more than $5$ steps, achieving comparable or even better performance to existing  $50$-step approaches on synthetic, real-world, and AIGC benchmarks.
	\end{compactitem}

	\vspace{-2mm}
	\section{Related Work}
	\vspace{-1mm}
	
	\noindent\textbf{Diffusion-based Video Restoration.}
	Diffusion models~\cite{rombach2022high,podell2023sdxl,blattmann2023stable,yang2024cogvideox} have substantially advanced video restoration. Early studies mainly focused on static image enhancement~\cite{wang2024exploiting,yu2024scaling,chen2025faithdiff}. While some methods augmenting 2D image backbones with auxiliary temporal modules~\cite{zhou2024upscale,yang2024motion} often struggle under severe degradations, recent Diffusion Transformers (DiT)~\cite{peebles2023scalable,yang2024cogvideox} exhibit superior spatiotemporal modeling. Consequently, DiT-based models like SeedVR~\cite{wang2025seedvr}, STAR~\cite{xie2025star}, and Vivid-VR~\cite{bai2025vivid} have achieved state-of-the-art detail recovery by introducing specialized architectures or objectives. However, their reliance on extensive iterative denoising (\eg, $50$ steps) incurs prohibitive computational latency, severely limiting practical deployment in real-world scenarios.
	
	\vspace{1mm}
	\noindent\textbf{Accelerated Diffusion for Video Restoration.}
	To accelerate inference, recent works~\cite{wang2025seedvr2,chen2025dove,zhuang2025flashvsr} aggressively distill models into single-step generators. Yet, by discarding progressive error correction, they often yield over-smoothed details and temporal flickering under complex spatiotemporal degradations. Alternatively, few-step image restoration paradigms~\cite{cui2024taming,yue2025arbitrary} jump-start the denoising from intermediate latents by bypassing early low-SNR timesteps. This naturally aligns with restoration tasks, where generating coarse structures from scratch is largely redundant. In particular, Cui \etal~\cite{cui2024taming} introduced a domain-shift strategy, allowing the denoising to jump-start from noised LQ inputs. Yue \etal~\cite{yue2025arbitrary} used an auxiliary predictor with frozen diffusion backbone to achieve few-step inference.
	However, extending this jump-starting paradigm to video mandates joint optimization of the predictor and denoiser to handle severe spatiotemporal degradations. Since naive joint training disrupts the diffusion trajectory and causes visual artifacts, our proposed SATB effectively resolves this train-inference discrepancy to enable robust few-step video restoration.

	\vspace{-2mm}
	\section{Method}
	\label{sec:method}
	\vspace{-1mm}
	
	Fig.~\ref{fig:framework} illustrates an overview of the proposed SATB-VR. In this section, we present the preliminaries, model architecture, and training strategy of the proposed method.

	\subsection{Preliminaries}
	\vspace{-1mm}
	
	In this work, we follow the jump-starting paradigm~\cite{yue2025arbitrary}. Given a selected inference timestep schedule $\{\tau_i\}_{i=1}^s$, the key lies in bypassing the previous timesteps and directly computing the latent $x_{\tau_s}$ at the starting timestep $\tau_s$. According to the standard forward diffusion process, the latent $x_{\tau_s}$ can be calculated by:
	\begin{equation}
		\setlength{\abovedisplayskip}{3pt}  
		\setlength{\belowdisplayskip}{3pt}  
		x_{\tau_s} = \sqrt{\bar{\alpha}_{\tau_s}} x_{gt} + \sqrt{1-\bar{\alpha}_{\tau_s}} \epsilon,~~ \epsilon \sim \mathcal{N}(0,I) ,
		\label{eq:x-ts-diffusion}
	\end{equation}
	where $x_{gt}$ is the ground-truth (GT) latent, $\bar{\alpha}_{\tau_s}$ denotes the cumulative noise schedule parameter, and $\epsilon$ is the noise.
	
	Since the clean $x_{gt}$ is inaccessible during inference in restoration tasks, \cite{yue2025arbitrary} approximates it using the low-quality (LQ) latent $x_{lq}$, and adopts an auxiliary predictor $\mathcal{F}_p$ to estimate the noise component and correct the offset. Consequently, the computation of $x_{\tau_s}$ is reformulated as:
	\begin{equation}
		\setlength{\abovedisplayskip}{3pt}  
		\setlength{\belowdisplayskip}{3pt}  
		x_{\tau_s} = \sqrt{\bar{\alpha}_{\tau_s}} x_{lq} + \sqrt{1-\bar{\alpha}_{\tau_s}} \mathcal{F}_p(x_{lq}, \tau_s).
		\label{eq:x-ts-invsr}
	\end{equation}
	Once $x_{\tau_s}$ is obtained, the pre-trained denoiser $\mathcal{F}_d$ iteratively denoises it over a few timesteps $\{\tau_i\}_{i=1}^s$ (\eg, $s \le 5$). Notably, \cite{yue2025arbitrary} optimizes only the predictor $\mathcal{F}_p$ during training, keeping the denoiser $\mathcal{F}_d$ strictly frozen.

	\begin{figure}[!t]
		\centering
		\includegraphics[width=0.99\linewidth]{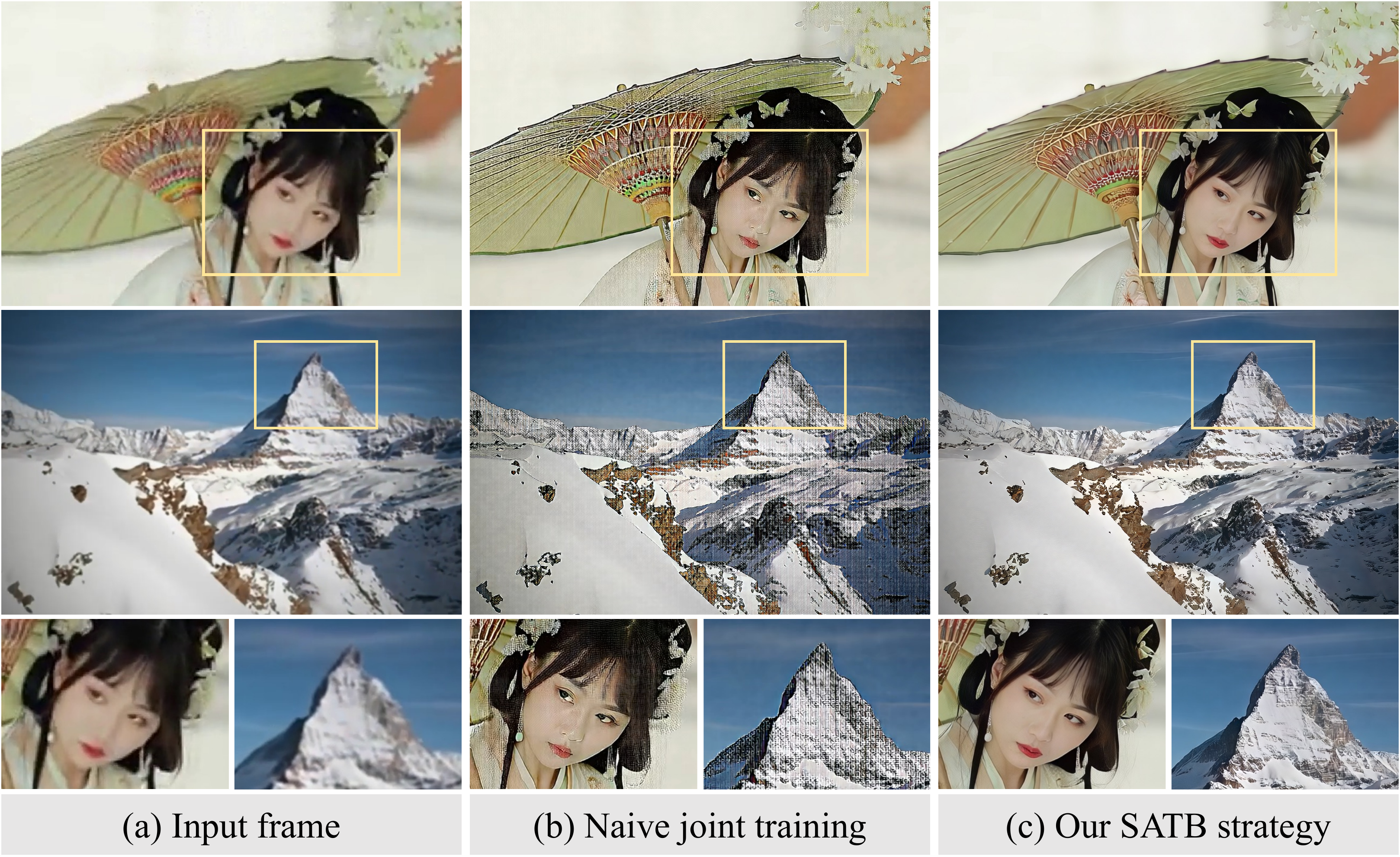}
		\vspace{1mm}
		\caption{%
			Effect of the proposed SATB strategy. Naive joint training suffers from a severe train-inference discrepancy, resulting in noticeable visual artifacts. In contrast, our proposed SATB strategy effectively addresses this issue, enabling robust joint training and yielding high-quality results.(\textbf{Zoom-in for best view})
		}
		\label{fig:visual-artifact}
	\end{figure}

	\subsection{Model Architectures}
	\vspace{-1mm}
	
	We leverage the pre-trained CogVideoX1.5-5B~\cite{yang2024cogvideox} as our base model, and extract text conditions from LQ inputs via CogVLM2-Video~\cite{yang2024cogvideox}. For brevity, the text condition is omitted in subsequent formulations.

	\vspace{1mm}
	\noindent\textbf{Auxiliary Predictor Design.} 
	In Eq.~\eqref{eq:x-ts-invsr}, $\mathcal{F}_p$ is burdened with both offset correction and noise estimation. Since the noise component $\epsilon$ strictly follows $\mathcal{N}(0,I)$, we explicitly resample it to relieve $\mathcal{F}_p$ of noise estimation, allowing it to focus entirely on offset correction. Accordingly, the computation of the latent $x_{\tau_s}$ is reformulated as:
	\begin{equation}
		\setlength{\abovedisplayskip}{3pt} 
		\setlength{\belowdisplayskip}{3pt} 
		x_{\tau_s} = \sqrt{\bar{\alpha}_{\tau_s}} x_p^{\tau_s} + \sqrt{1-\bar{\alpha}_{\tau_s}} \epsilon, \quad \epsilon \sim \mathcal{N}(0,I),
		\label{eq:x-ts-ours}
	\end{equation}
	where $x_p^{\tau_s} = \mathcal{F}_p(x_{lq}, \tau_s)$ denotes the predicted offset-corrected latent.
	We initialize $\mathcal{F}_p$ with pre-trained Expert Transformer blocks from the base model and optimize it efficiently via LoRA~\cite{hu2022lora} fine-tuning technique.

	\vspace{1mm}
	\noindent\textbf{Conditional Video Denoiser.} 
	Relying on a frozen denoiser~\cite{yue2025arbitrary} limits the model's capacity to adaptively correct the predictor's errors and fails to incorporate the LQ video as a necessary control condition. These flaws are often more pronounced for the video restoration task when handling complex spatiotemporal degradations.
	Thus, we redesign the denoiser by integrating a ControlNet branch~\cite{bai2025vivid} for spatial-temporal LQ condition injection. Furthermore, we co-fine-tune the base denoiser using LoRA, enabling it to dynamically compensate for the predictor's errors.

	\vspace{1mm}
	\noindent\textbf{Few-Step Inference Pipeline.} 
	Our framework inherently supports flexible few-step inference. Specifically, the denoising starting point $\tau_s$ is flexibly selected based on the desired inference steps $s$. A larger $s$ allows for a higher $\tau_s$, thereby unlocking stronger generative capacity for fine detail recovery. Once $\tau_s$ is determined, the complete schedule $\{\tau_i\}_{i=1}^s$ is derived via uniform sampling across the remaining timesteps. This flexibility facilitates a controllable trade-off between restoration quality and efficiency.

	\subsection{Training Strategy}
	\vspace{-1mm}

	\vspace{1mm}
	\noindent\textbf{SNR-Aware Trajectory Blending Strategy.}
	In standard diffusion training, the forward process adds noise to the GT latent $x_{gt}$. However, during inference, our denoiser initiates from the predictor's approximated output $x_p^{\tau_s}$. Optimizing the denoiser solely on the standard GT trajectory leaves it unable to perceive and correct the predictor's estimation errors. Conversely, directly basing the forward process on $x_p^{\tau_s}$ across all timesteps disrupts the native diffusion manifold. This leads to a train-inference discrepancy, inevitably resulting in severe visual artifacts, as shown in Fig.~\ref{fig:visual-artifact}(b).
	
	To resolve this discrepancy, the forward process must seamlessly bridge both states: anchoring the initial state to $x_p^{\tau_s}$ at $\tau_s$, while smoothly transitioning towards the clean GT trajectory at subsequent steps. 
	However, employing heuristic linear/cosine interpolations risks pushing the latents off the true diffusion manifold, as they are agnostic to the underlying noise schedule.
	We note that in the diffusion process, the attenuation of initial prediction errors inherently scales with the relative SNR (detailed derivations in Appendix~\ref{sec:appendix-derivation-satb}).
	Motivated by this, we propose the SNR-Aware Trajectory Blending (SATB) strategy.
	Given sampled timesteps $\tau_s \sim U\{1,\dots,T\}$ and $t \sim U\{1,\dots,\tau_s\}$, SATB constructs a blended latent $x_b^t$ via relative SNR modulation. This physically aligns the initial prediction error with the noise schedule of the denoiser at step $t$:
	\begin{equation}
		\setlength{\abovedisplayskip}{3pt}
		\setlength{\belowdisplayskip}{3pt}
		x_b^t = x_{gt} + \frac{\gamma_{\tau_s}}{\gamma_t}(x_p^{\tau_s} - x_{gt}),
		\label{eq:satb-blend}
	\end{equation}
	where $U\{\cdot\}$ denotes the uniform distribution; $T$ denotes the total denoising steps; $\gamma_{\tau_s}$ and $\gamma_t$ are the SNRs at $\tau_s$ and $t$, respectively. 
	This dynamic blending ensures that at $t = \tau_s$, the denoiser is forced to compensate for the predictor's bias. As generation proceeds and SNR increases, the scaling factor $\frac{\gamma_{\tau_s}}{\gamma_t}$ decays, smoothly guiding the trajectory to converge onto the clean $x_{gt}$ manifold without corrupting the underlying marginal distribution. Consequently, SATB eliminates the train-inference discrepancy, enabling robust joint training and high-quality iterative inference (Fig.~\ref{fig:visual-artifact}(c)).
	
	Subsequently, standard noise-addition is applied to $x_b^t$ to produce the noisy input for the denoiser $\mathcal{F}_d$. We optimize the network via the $v$-prediction objective:
	\begin{equation}
		\setlength{\abovedisplayskip}{3pt}  
		\setlength{\belowdisplayskip}{3pt}  
		\mathcal{L}_{diff} = \mathbb{E} \left [ \left \| v - \mathcal{F}_d( \mathbb{N}(x_b^t, t, \epsilon ), x_{lq}, t ) \right \|_2^2  \right ],
		\label{eq:diffusion-loss}
	\end{equation}
	where $\mathbb{N}(\cdot, t, \epsilon)$ is the forward noise-addition operation defined in Eq.~\eqref{eq:x-ts-diffusion} with $\epsilon \sim \mathcal{N}(0,I)$, and the optimization target is derived from $x_{gt}$ as $v = \sqrt{\bar{\alpha}_t}\epsilon - \sqrt{1-\bar{\alpha}_t}x_{gt}$.

	
\begin{table*}[!t]
\caption{Quantitative comparisons on benchmarks, including synthetic (SPMCS~\cite{tao2017detail}, UDM10~\cite{yi2019progressive}, YouHQ40~\cite{zhou2024upscale}), real-world (VideoLQ~\cite{chan2022investigating}, UGC50~\cite{bai2025vivid}), and AIGC (AIGC50~\cite{bai2025vivid}) videos. The best and second performances are marked in {\color[HTML]{F94848} red} and {\color[HTML]{3166FF} blue}, respectively.}
\vspace{1mm}
\renewcommand\arraystretch{1.1}
\footnotesize
\begin{tabular}{c|c|cccc|ccc|cc}
\toprule
Datasets    & Metrics    & \makecell{~~~~UAV~~~~ \\ (\textit{30 steps})}    & \makecell{STAR \\ (\textit{15 steps})}    & \makecell{SeedVR \\ (\textit{50 steps})}    & \makecell{~~Vivid-VR~~ \\ (\textit{50 steps})}    & \makecell{~~~DOVE~~~ \\ (\textit{1 step})}    & \makecell{SeedVR2 \\ (\textit{1 step})}    & \makecell{FlashVSR \\ (\textit{1 step})}    & \makecell{~~~Ours~~~ \\ (\textit{1 step})}    & \makecell{~~~Ours~~~ \\ (\textit{5 steps})}    \\ \hline
           & PSNR $\uparrow$     & 23.01 & 24.18 & 24.08 & 21.73  & {\color[HTML]{3166FF} 24.80}     & {\color[HTML]{F94848} 26.07} & 23.44 & 24.18 & 21.69     \\
           & SSIM $\uparrow$     & 0.606 & 0.720 & 0.689 & 0.604  & {\color[HTML]{3166FF} 0.754}     & {\color[HTML]{F94848} 0.777}     & 0.670 & 0.707 & 0.599     \\
           & LPIPS $\downarrow$  & 0.277     & 0.301 & 0.263 & 0.278  & {\color[HTML]{F94848} 0.168} & {\color[HTML]{3166FF} 0.191} & 0.226 & 0.197& 0.294     \\ \cline{2-11} 
           & MANIQA $\uparrow$   & 0.385     & 0.229 & 0.315 & {\color[HTML]{3166FF} 0.410}  & 0.346     & 0.305     & 0.381 & 0.384 & {\color[HTML]{F94848} 0.433} \\
           & MUSIQ $\uparrow$    & 66.11 & 30.62 & 56.99 & {\color[HTML]{3166FF} 70.03}  & 63.29     & 53.23     & 67.91 & 67.82 & {\color[HTML]{F94848} 72.14} \\
           & CLIP-IQA $\uparrow$ & 0.427     & 0.254 & 0.347 & 0.483  & 0.410     & 0.325     & {\color[HTML]{3166FF} 0.571} & 0.514 & {\color[HTML]{F94848} 0.625} \\
\multirow{-7}{*}{SPMCS}  & DOVER $\uparrow$    & 8.987     & 4.266 & 9.779 & {\color[HTML]{3166FF} 11.35}  & 9.898     & 8.625     & 10.38 & 10.65 & {\color[HTML]{F94848} 11.93} \\ \hline
           & PSNR $\uparrow$     & 28.20 & 27.29 & 27.80 & 24.54  & {\color[HTML]{F94848} 30.53}     & {\color[HTML]{3166FF} 29.04}     & 26.36 & 28.67 & 25.66     \\
           & SSIM $\uparrow$     & 0.826     & 0.855 & 0.848 & 0.761  & {\color[HTML]{F94848} 0.894}     & {\color[HTML]{3166FF} 0.884}     & 0.797 & 0.859 & 0.772     \\
           & LPIPS $\downarrow$  & 0.196     & 0.167 & 0.148 & 0.243  & {\color[HTML]{F94848} 0.101}     & {\color[HTML]{3166FF} 0.117} & 0.182 & 0.150 & 0.229     \\ \cline{2-11} 
           & MANIQA $\uparrow$   & 0.297     & 0.260 & 0.264 & 0.359  & 0.296     & 0.262     & 0.364 & {\color[HTML]{3166FF} 0.381} & {\color[HTML]{F94848} 0.416} \\
           & MUSIQ $\uparrow$    & 56.19     & 45.38 & 50.29 & 64.71  & 55.17     & 48.91     & 65.07 & {\color[HTML]{3166FF} 65.83} & {\color[HTML]{F94848} 69.75} \\
           & CLIP-IQA $\uparrow$ & 0.333     & 0.289 & 0.273 & 0.426  & 0.340     & 0.272     & {\color[HTML]{3166FF} 0.556} & 0.507 & {\color[HTML]{F94848} 0.601} \\
\multirow{-7}{*}{UDM10}       & DOVER $\uparrow$    & 9.774     & 9.454 & 9.349 & {\color[HTML]{3166FF} 11.97}  & 10.41     & 8.752     & 11.60 & 10.98 & {\color[HTML]{F94848} 12.49} \\ \hline
           & PSNR $\uparrow$     & 22.31     & 22.92 & 22.46 & 21.31  & {\color[HTML]{F94848} 24.10}     & {\color[HTML]{3166FF} 24.00} & 22.56 & 23.67 & 21.98     \\
           & SSIM $\uparrow$     & 0.592     & 0.657 & 0.621 & 0.579  & {\color[HTML]{3166FF} 0.688}    & {\color[HTML]{F94848} 0.693} & 0.602 & 0.657 & 0.589     \\
           & LPIPS $\downarrow$  & 0.340     & 0.433 & {\color[HTML]{3166FF} 0.240} & 0.357  & 0.283 & {\color[HTML]{F94848} 0.185} & 0.290 & 0.281 & 0.303     \\ \cline{2-11} 
           & MANIQA $\uparrow$   & 0.344     & 0.240 & 0.315 & 0.372  & 0.304     & 0.314     & {\color[HTML]{3166FF} 0.367} & 0.354 & {\color[HTML]{F94848} 0.396} \\
           & MUSIQ $\uparrow$    & 65.97     & 36.36 & 61.91 & {\color[HTML]{3166FF} 70.55}  & 60.65     & 59.34     & 69.62 & 67.91 & {\color[HTML]{F94848} 72.34} \\
           & CLIP-IQA $\uparrow$ & 0.427     & 0.279 & 0.360 & 0.447  & 0.356     & 0.336     & {\color[HTML]{3166FF} 0.590} & 0.486 & {\color[HTML]{F94848} 0.603}     \\
\multirow{-7}{*}{YouHQ40}     & DOVER $\uparrow$    & 12.36     & 7.868 & 14.00 & {\color[HTML]{F94848} 14.61}  & 12.52 & 12.80     & 13.84 & 13.25 & {\color[HTML]{3166FF} 14.46} \\ \hline
           & MANIQA $\uparrow$   & 0.275     & 0.271 & 0.223 & 0.319  & 0.272     & 0.221     & 0.299 & {\color[HTML]{3166FF} 0.356} & {\color[HTML]{F94848} 0.383} \\
           & MUSIQ $\uparrow$    & 55.82     & 50.52 & 46.49 & 62.47  & 55.11     & 43.41     & 61.88 & {\color[HTML]{3166FF} 65.59} & {\color[HTML]{F94848} 68.28} \\
           & CLIP-IQA $\uparrow$ & 0.262     & 0.265 & 0.229 & 0.338  & 0.271     & 0.220     & 0.405 & {\color[HTML]{3166FF} 0.436} & {\color[HTML]{F94848} 0.485} \\
\multirow{-4}{*}{VideoLQ}     & DOVER $\uparrow$    & 7.777     & 8.758 & 7.240 & {\color[HTML]{3166FF} 9.743}  & 8.780     & 6.331     & 9.363 & 9.577 & {\color[HTML]{F94848} 9.930} \\ \hline
           & MANIQA $\uparrow$   & 0.314     & 0.325 & 0.262 & 0.376  & 0.320     & 0.253     & 0.372 & {\color[HTML]{3166FF} 0.402} & {\color[HTML]{F94848} 0.430} \\
           & MUSIQ $\uparrow$    & 54.71     & 55.01 & 49.76 & 67.61  & 57.82     & 46.12     & 65.66 & {\color[HTML]{3166FF} 68.52} & {\color[HTML]{F94848} 71.55} \\
           & CLIP-IQA $\uparrow$ & 0.353     & 0.353 & 0.305 & 0.450  & 0.353     & 0.276     & 0.563 & {\color[HTML]{3166FF} 0.571} & {\color[HTML]{F94848} 0.661} \\
\multirow{-4}{*}{UGC50}       & DOVER $\uparrow$    & 10.44     & 10.92 & 10.47 & {\color[HTML]{F94848} 14.46}  & 11.84     & 8.209     & 13.29 & 13.40 & {\color[HTML]{3166FF} 14.37} \\ \hline
           & MANIQA $\uparrow$   & 0.333     & 0.347 & 0.286 & 0.369  & 0.334     & 0.281     & 0.369 & {\color[HTML]{3166FF} 0.378} & {\color[HTML]{F94848} 0.415} \\
           & MUSIQ $\uparrow$    & 57.62     & 51.66  & 61.61 & {\color[HTML]{3166FF} 67.18}     & 62.07 & 49.35     & 66.60 & 66.08 & {\color[HTML]{F94848} 70.60} \\
           & CLIP-IQA $\uparrow$ & 0.376     & 0.309  & 0.378 & 0.445     & 0.379 & 0.290     & {\color[HTML]{3166FF} 0.579} & 0.493 & {\color[HTML]{F94848} 0.594} \\
\multirow{-4}{*}{AIGC50}      & DOVER $\uparrow$    & 12.28     & 12.10  & 14.46 & {\color[HTML]{3166FF} 14.51}     & 14.49 & 11.34     & 14.41 & 14.43 & {\color[HTML]{F94848} 15.14} \\
\bottomrule
\end{tabular}
\label{tab:quantitative-results}
\end{table*}

	\begin{figure*}[!t]
		\centering
		\includegraphics[width=1.0\linewidth]{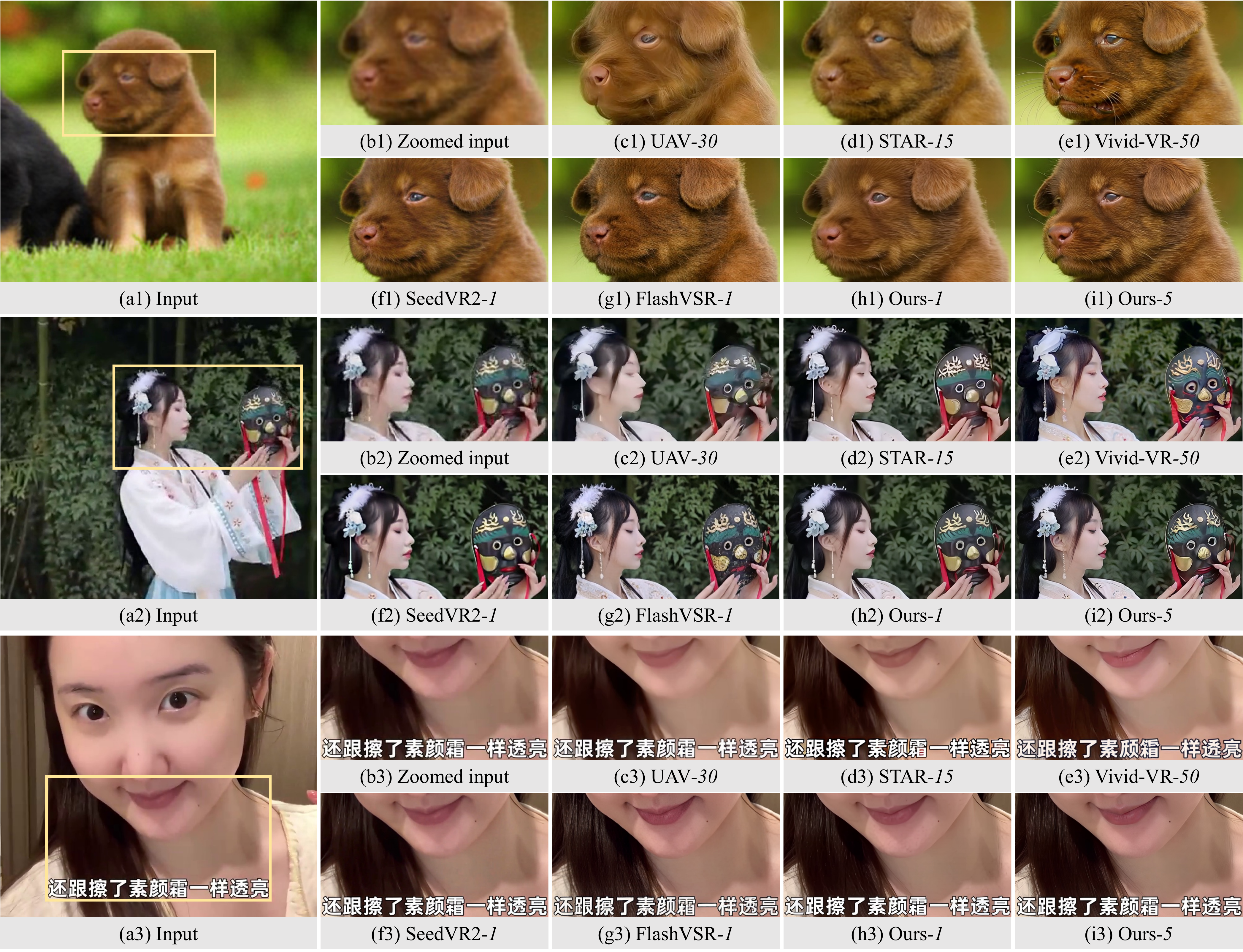}
		\caption{%
			Qualitative comparison results on synthetic ($1st$ row) and real-world ($2nd$ and $3rd$ rows) videos. The proposed method produces frames with strict structural fidelity, highly realistic textures, and notably better text rendering. (\textbf{Zoom-in for best view})
		}
		\label{fig:visual-compare}
	\end{figure*}
	
	\begin{figure}[!t]
		\centering
		\includegraphics[width=0.99\linewidth]{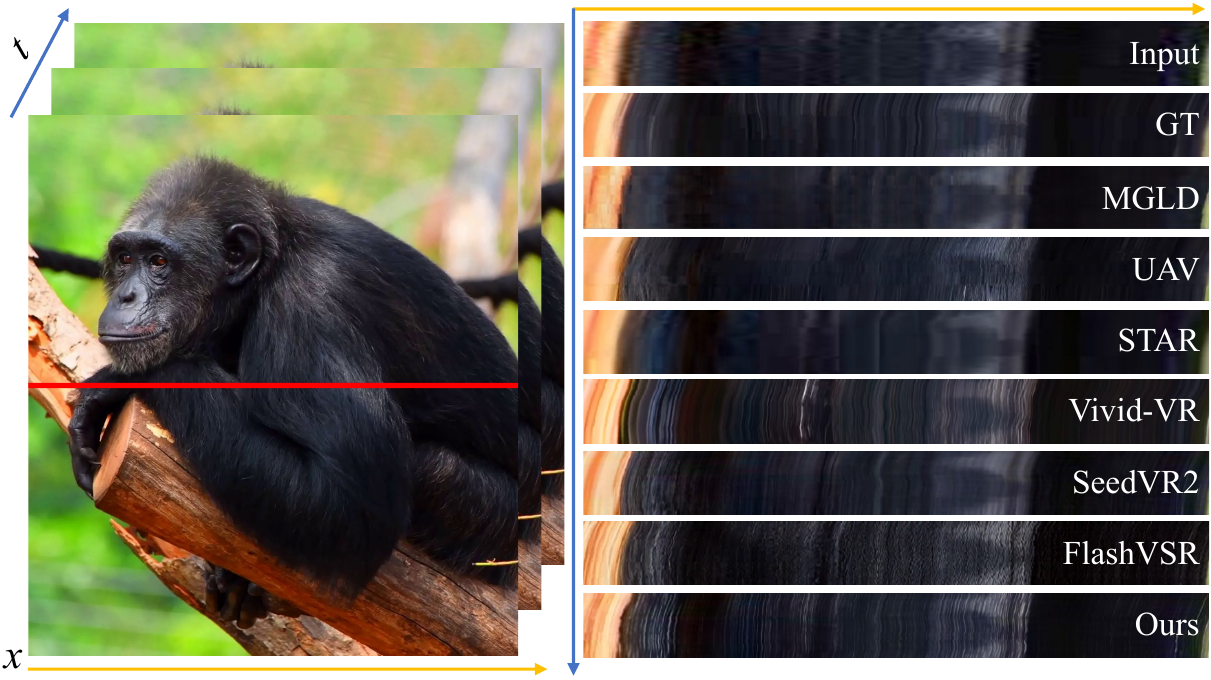}
		\caption{%
			Temporal profiles generated by stacking the {\color{red} red} line across frames. Unlike the severe temporal flickering in FlashVSR, our SATB-VR maintains better temporal consistency.
		}
		\label{fig:temporal-profile}
	\end{figure}

	\vspace{1mm}
	\noindent\textbf{Denoiser-Driven Consistency Loss.}
	To explicitly constrain the predictor, we propose the Denoiser-Driven Consistency (DDC) loss. Our goal is to align the deep representations elicited by the predictor's output $x_p^{\tau_s}$ with those of the ground-truth $x_{gt}$. By employing the denoiser $\mathcal{F}_d$ as a dynamic feature extractor, DDC minimizes the Mean Squared Error (MSE) between their intermediate activations:
	\begin{equation}
		\setlength{\abovedisplayskip}{3pt}  
		\setlength{\belowdisplayskip}{3pt}  
		\mathcal{L}_{ddc} = \mathbb{E} \left [ \left \| \Phi(\hat{x}_p^{\tau_s}, x_{lq}, \tau_s) - \Phi(\hat{x}_{gt} , x_{lq}, \tau_s) \right \|_2^2 \right ],
		\label{eq:ddc-loss}
	\end{equation}
	where $\hat{x}_p^{\tau_s} = \mathbb{N}(x_p^{\tau_s}, \tau_s, \epsilon)$ and $\hat{x}_{gt} = \mathbb{N}(x_{gt}, \tau_s, \epsilon)$ denote the respective noisy latents, and $\Phi(\cdot)$ represents the intermediate feature activations extracted from the denoiser $\mathcal{F}_d$, which is detached during backpropagation. 
	Crucially, by utilizing the dynamically updating denoiser, DDC provides the predictor with up-to-date gradients, forcing $x_p^{\tau_s}$ to continuously align with the evolving generative manifold.
	
	Finally, the overall loss is formulated as:
	\begin{equation}
		\setlength{\abovedisplayskip}{3pt}  
		\setlength{\belowdisplayskip}{3pt}  
		\mathcal{L}_{total} = \mathcal{L}_{diff} + \lambda \mathcal{L}_{ddc},
		\label{eq:total-loss}
	\end{equation}
	where $\lambda$ is a balancing weight hyperparameter.

	\vspace{-2mm}
	\section{Experimental Results}
	\label{sec:experimental-results}
	\vspace{-1mm}
	
	In this section, we evaluate the proposed SATB-VR on synthetic, real-world, and AIGC benchmarks and compare it with state-of-the-art methods.

	\vspace{-1mm}
	\subsection{Implementation Details}
	\label{sec:implementation-details}
	\vspace{-1mm}
	
	We curate our training dataset from OpenVid-1M~\cite{nan2025openvid}, ShareGPT4Video~\cite{chen2024sharegpt4video}, and InternVid~\cite{wang2024internvid}. To ensure high visual quality, videos are strictly filtered by resolution, duration, no scene transitions, and high no-reference score~\cite{zhang2023md}. By employing CogVLM2-Video~\cite{yang2024cogvideox} for text annotation and the degradation pipeline from~\cite{wang2021real} for LQ synthesis, we construct approximately $200K$ HQ-LQ video-text pairs.
	During training, we incorporate the concept distillation strategy~\cite{bai2025vivid} to mitigate distribution drift and set the DDC loss weight to $\lambda = 1.0$. SATB-VR is optimized for $20K$ iterations on $16$ NVIDIA H20 GPUs (batch size $1$ per GPU) using the AdamW optimizer~\cite{loshchilov2017decoupled} with an initial learning rate of $10^{-4}$ and cosine annealing~\cite{wang2019edvr}. For inference, we employ the DPM-Solver~\cite{lu2025dpm} with default settings of $s = 5$ and $\tau_s = 399$. Configurations for other inference steps are detailed in Appendix~\ref{sec:appendix-more-implementation-details}.

	\vspace{-1mm}
	\subsection{Quantitative Results}
	\label{sec:quantitative-results}
	\vspace{-1mm}

	We evaluate SATB-VR against state-of-the-art methods, including multi-step video restoration (UAV~\cite{zhou2024upscale}, STAR~\cite{xie2025star}, SeedVR~\cite{wang2025seedvr}, Vivid-VR~\cite{bai2025vivid}), and aggressive single-step methods (DOVE~\cite{chen2025dove}, SeedVR2~\cite{wang2025seedvr2}, FlashVSR~\cite{zhuang2025flashvsr}). 
	The comparisons are conducted across synthetic (SPMCS~\cite{tao2017detail}, UDM10~\cite{yi2019progressive}, YouHQ40~\cite{zhou2024upscale}) and real-world/AIGC benchmarks (VideoLQ~\cite{chan2022investigating}, UGC50~\cite{bai2025vivid}, AIGC50~\cite{bai2025vivid}). For real-world and AIGC scenarios lacking GT, we use no-reference quality metrics (MANIQA~\cite{yang2022maniqa}, MUSIQ~\cite{ke2021musiq}, CLIP-IQA~\cite{wang2023exploring}, DOVER~\cite{wu2023exploring}). For synthetic datasets, we supplemented the evaluation with full-reference metrics (PSNR, SSIM, and LPIPS~\cite{zhang2018unreasonable}).

	\begin{figure*}[!t]
		\centering
		\includegraphics[width=1.0\linewidth]{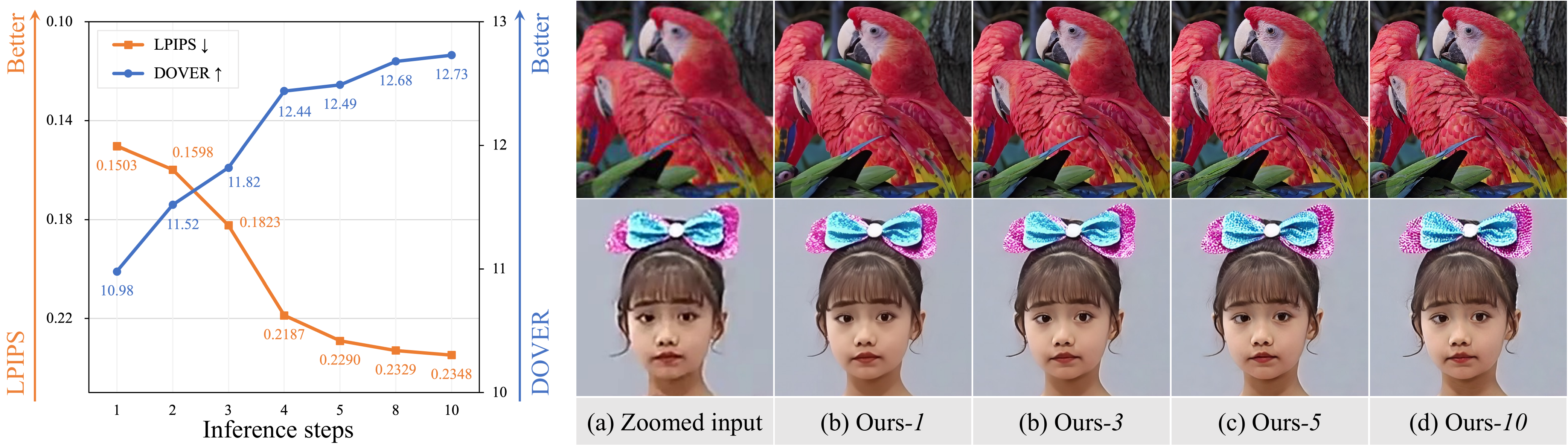}
		\caption{%
			Performance trends at various inference steps, where the y-axis for LPIPS is inverted for better visualization (i.e., upward indicates better performance). Notably, as inference steps increase, the perceptual DOVER score improves, while LPIPS incurs a penalty.
			The visual patches in (b)$\sim$(d) show that while $1$-step outputs yield better LPIPS scores due to their conservative and smooth nature, the $5/10$-step outputs synthesize much richer and realistic textures that align better with human perception.
		}
		\label{fig:inference-steps}
	\end{figure*}

\begin{table*}[!t]
\caption{Ablation studies on the UGC50~\cite{bai2025vivid} testset, where ``LQ Cond. Denoiser" indicates incorporating the LQ video as a control condition, ``Denoiser FT" denotes whether the denoiser is co-optimized. For the blending mechanism, we compare SATB with no blending (``\ding{55}"), ``\textit{Linear Blend}", and ``\textit{Cosine Blend}" in Eq.~\eqref{eq:satb-blend}. (h) represents our full SATB-VR model. Best results are highlighted in \textbf{bold}.}
\renewcommand\arraystretch{1.1}
\footnotesize
\begin{tabular}{c|c|c|c|c|c|cccc}
\toprule
\multirow{2}{*}{Methods} & \multicolumn{2}{c|}{{Model Architectures}} & \multicolumn{3}{c|}{Training Strategy} & \multirow{2}{*}{MANIQA$\uparrow$} & \multirow{2}{*}{MUSIQ$\uparrow$} & \multirow{2}{*}{CLIP-IQA$\uparrow$} & \multirow{2}{*}{DOVER$\uparrow$} \\ \cline{2-6}
        &  Predictor   &  LQ Cond. Denoiser      & Denoiser FT    & SATB    & DDC Loss   &      &       &          &       \\ \hline
(a)     & \Checkmark  & \ding{55}        & \ding{55}             & \ding{55}       & \Checkmark        & 0.367    & 66.83    & 0.561    & 13.87    \\
(b)     & \Checkmark  & \ding{55}        & \Checkmark             & \Checkmark       & \Checkmark        & 0.396    & 68.88    & 0.611    & 14.26    \\
(c)     & \Checkmark  & \Checkmark        & \Checkmark             & \ding{55}       & \Checkmark        & 0.380    & 52.32    & 0.523    & 7.283    \\
(d)     & \Checkmark  & \Checkmark        & \Checkmark             & \textit{Linear Blend}       & \Checkmark        & 0.414    & 69.60    & 0.644    & 13.36    \\
(e)     & \Checkmark  & \Checkmark        & \Checkmark             & \textit{Cosine Blend}       & \Checkmark        & 0.419    & 69.47    & 0.652    & 12.97    \\
(f)     & \ding{55}  & \Checkmark        & \Checkmark             & \Checkmark       & \ding{55}        & 0.405    & 69.33    & 0.623    & 13.75    \\
(g)     & \Checkmark  & \Checkmark        & \Checkmark             & \Checkmark       & \ding{55}        & 0.401    & 68.95    & 0.567    & 13.80    \\
(h)     & \Checkmark  & \Checkmark        & \Checkmark             & \Checkmark       & \Checkmark        & \textbf{0.430}    & \textbf{71.55}    & \textbf{0.661}    & \textbf{14.37}    \\
\bottomrule
\end{tabular}
\label{tab:ablation-study}
\end{table*}

	Tab.~\ref{tab:quantitative-results} summarizes the quantitative evaluations across all benchmarks. SATB-VR achieves state-of-the-art performance across nearly all no-reference metrics.
	While single-step methods (e.g., DOVE~\cite{chen2025dove}) yield competitive full-reference scores, it is a well-established phenomenon that such metrics inherently favor conservative, over-smoothed predictions over realistic generative textures~\cite{blau2018perception,yu2024scaling}.
	As illustrated in Fig.~\ref{fig:inference-steps}, while increasing inference steps enriches high-frequency details and improves perceptual quality (\eg, DOVER), it simultaneously incurs slight spatial deviations that are heavily penalized by LPIPS.
	Given that the primary objective of generative video restoration is to synthesize visually realistic and pleasing contents, no-reference metrics provide a much more accurate reflection of human preference. On these perceptual metrics, our SATB-VR achieves comparable or superior quality to standard diffusion frameworks requiring $50$ iterations (\eg, Vivid-VR~\cite{bai2025vivid}), demonstrating an exceptional balance between generation quality and inference efficiency.

	\vspace{-1mm}
	\subsection{Qualitative Results}
	\label{sec:qualitative-results}
	\vspace{-1mm}

	Figs.~\ref{fig:teaser} and~\ref{fig:visual-compare} present qualitative comparisons on synthetic and real-world videos.
	While $50$-step methods (e.g., Vivid-VR~\cite{bai2025vivid}) exhibit decent texture realism, their excessive generative capacity often leads to severe hallucination issues, compromising structural fidelity. As illustrated in Fig.~\ref{fig:visual-compare}, this manifests as noticeable structural deviations (e1), incorrect material rendering (e2), and garbled text (e3). Conversely, our SATB-VR effectively constrains this generative drift, ensuring strict consistency with the input.
	Meanwhile, lacking iterative correction, aggressive single-step approaches struggle to resolve complex degradations, typically yielding over-smoothed results or unnatural artifact textures. Notably, FlashVSR~\cite{zhuang2025flashvsr} frequently suffers from chaotic noise patterns (see Fig.~\ref{fig:teaser}(d), Fig.~\ref{fig:visual-compare}(g1)-(g3)) and irregular temporal flickering (see Fig.~\ref{fig:temporal-profile}, our supplementary video in Appendix~\ref{sec:appendix-more-visualization-results}), which severely impairs the perceptual viewing experience. In contrast, SATB-VR achieves highly realistic textures and better temporal consistency.

	\vspace{-4mm}
	\section{Analysis and Discussions}
	\label{sec:analysis-and-discussion}
	\vspace{-1mm}
	
	We have demonstrated that the proposed SATB-VR performs favorably against state-of-the-art methods. In this section, we perform further analysis on the key components and discuss the limitations of our current method.

	\vspace{-1mm}
	\subsection{Effect of the Conditional Denoiser}
	\label{sec:effect-conditional-denoiser}
	\vspace{-1mm}
	
\begin{figure}[!t]
		\centering
		\includegraphics[width=0.99\linewidth]{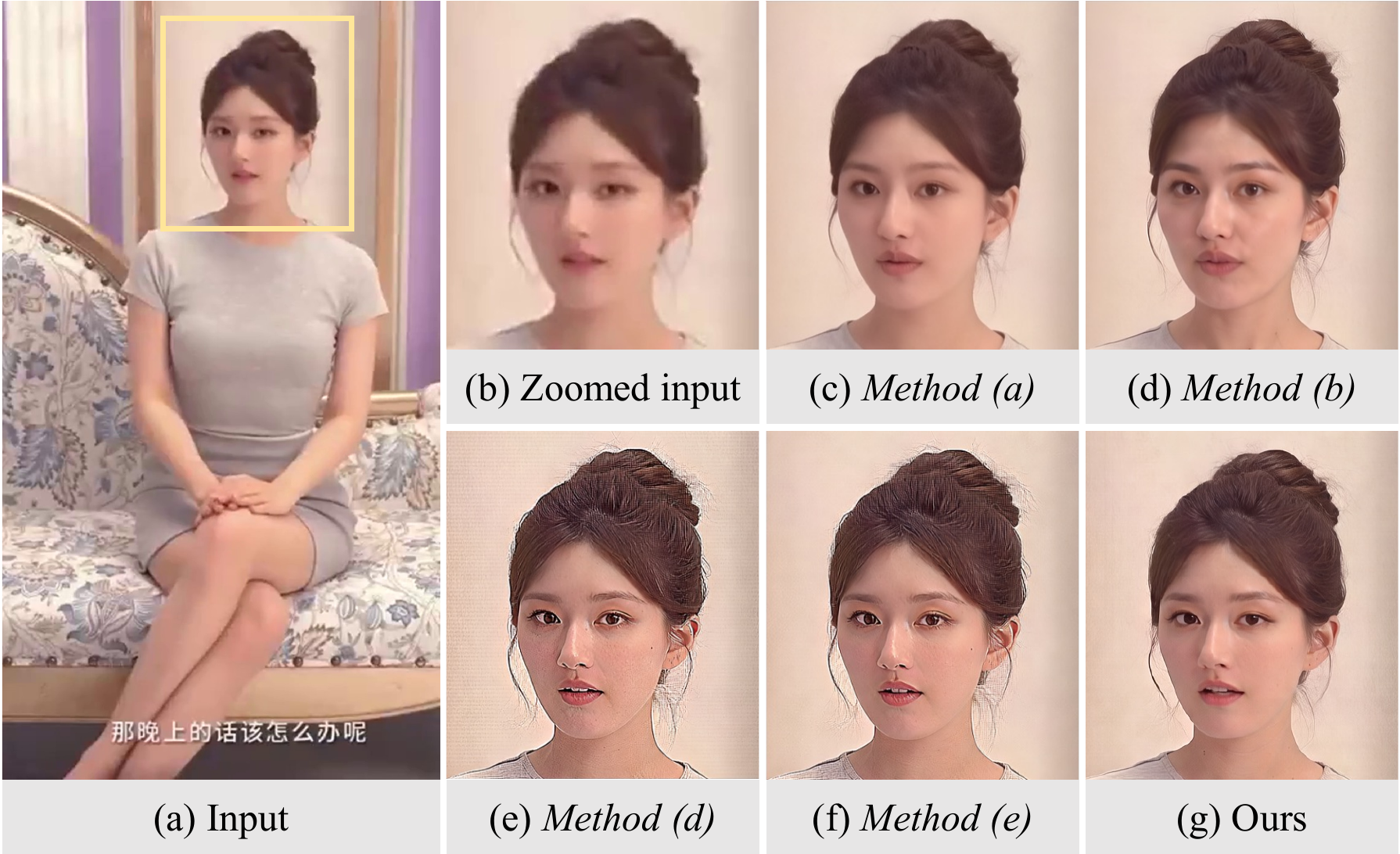}
		\vspace{1mm}
		\caption{%
			Visual comparisons of the ablation studies. The notation ``\textit{Method ($\cdot$)}'' corresponds to the respective configuration in Tab.~\ref{tab:ablation-study}. Compared to the baselines, our method effectively removes degradations while preserving faithful identities and expressions.
		}
		\label{fig:ablation-satb}
	\end{figure}

	Previous method~\cite{yue2025arbitrary} optimizes the predictor while keeping the denoiser frozen. As discussed, this design cannot adaptively compensate for predictor errors or incorporate explicit LQ conditioning, restricting its capacity for complex spatiotemporal degradations.
	We evaluate two baselines: (a) freezing the denoiser to mimic~\cite{yue2025arbitrary} (thus disabling LQ conditioning and SATB), and (b) co-optimizing both modules but omits the LQ control branch.
	Both baselines yield suboptimal quantitative results (Tab.~\ref{tab:ablation-study}(a)\&(b)). Visually (Fig.~\ref{fig:ablation-satb}(c)\&(d)), lacking explicit LQ conditioning for structural guidance, they struggle to remove complex degradations and suffer severe fidelity loss, causing identity and expression shifts. This verifies the indispensability of joint-training the denoiser with explicit LQ conditioning.

	\begin{figure}[!t]
		\centering
		\includegraphics[width=0.99\linewidth]{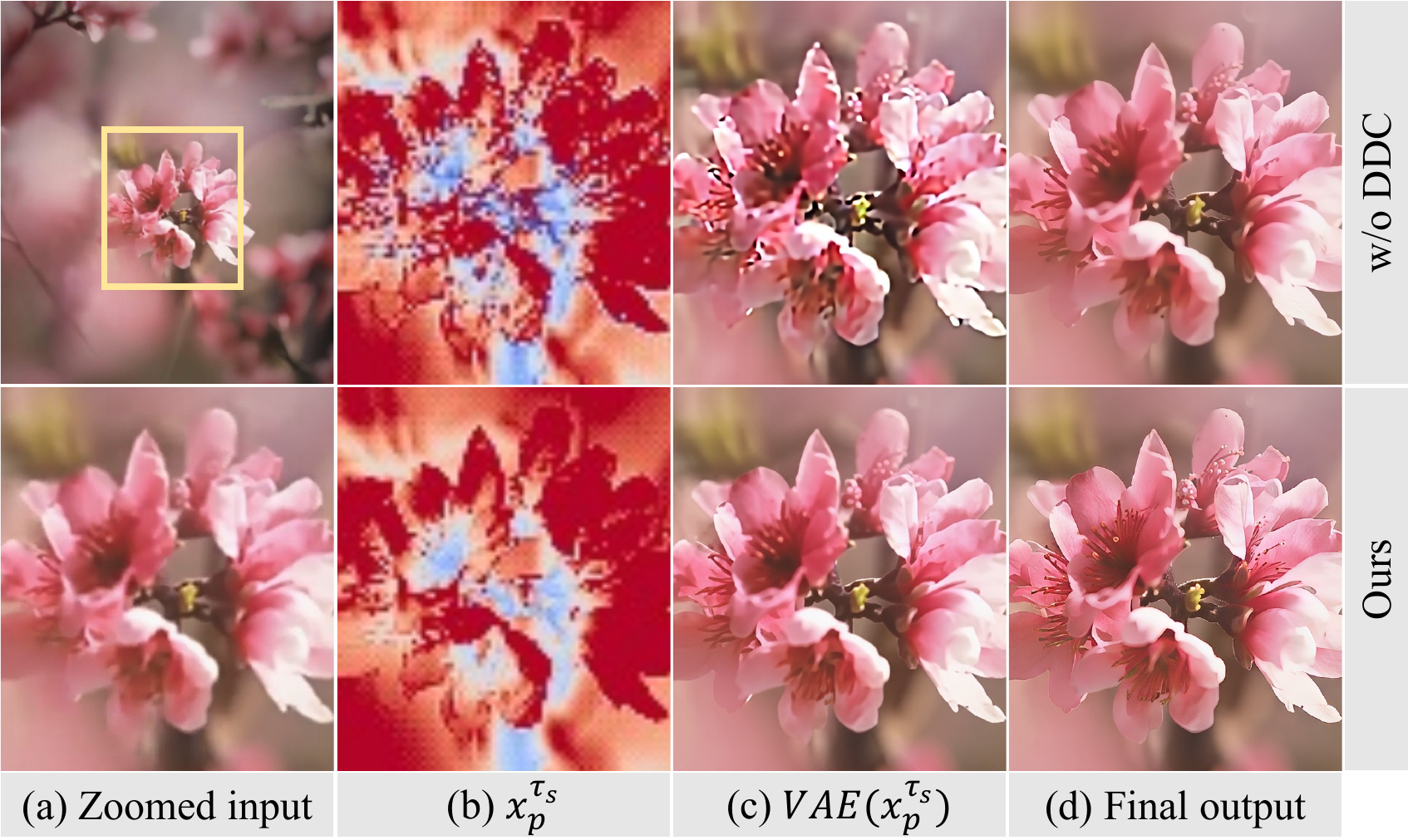}
		\vspace{1mm}
		\caption{%
			Visualizing the impact of the DDC loss. Columns show the zoomed input, feature heatmaps of the predicted latent $x_p^{\tau_s}$, its VAE-decoded spatial visualization, and the final restored output. Without DDC, $x_p^{\tau_s}$ suffers from noisy outliers, causing severe artifacts in the decoded space and a blurry final result.
		}
		\label{fig:ablation-ddc}
	\end{figure}

	\vspace{-1mm}
	\subsection{Effect of the SATB Strategy}
	\label{sec:effect-satb-strategy}
	\vspace{-1mm}
	
	Naive joint training, which directly feeds the noised predictor's output into the denoiser, inherently introduces a severe train-inference discrepancy. To explicitly verify this, we disable our SATB strategy and retrain this baseline method. As shown in Tab.~\ref{tab:ablation-study}(c), this straightforward approach causes a significant degradation in quantitative metrics. As corroborated by Fig.~\ref{fig:visual-artifact}(b), such manifold deviation leads to severe visual artifacts.
	
	Furthermore, we investigate our trajectory blending design. While recent work~\cite{cui2024taming} explores similar interpolation to bypass early diffusion steps, it relies on a timestep-based cosine function and initiates directly from the LQ latent. To validate our SNR-aware formulation, we replace our SNR coefficient in Eq.~\eqref{eq:satb-blend} with linear or cosine functions. As shown in Tab.~\ref{tab:ablation-study}(d)\&(e) and Fig.~\ref{fig:ablation-satb}(e)\&(f), both alternative schedules yield suboptimal results with visual artifacts. This confirms that anchoring the blending to the physical SNR is crucial for maintaining a continuous data manifold.
	Additionally, starting the trajectory directly from the raw LQ latent without the predictor significantly degrades performance (Tab.~\ref{tab:ablation-study}(f) vs (h)). Without explicit offset correction, complex degradation features irreversibly bleed into the generation process, leading to suboptimal results.

	\vspace{-1mm}
	\subsection{Effect of the DDC Loss}
	\label{sec:effect-ddc-loss}
	\vspace{-2mm}
	
	Relying solely on the global diffusion loss (i.e., $\mathcal{L}_{diff}$) provides only indirect supervision for the predictor. Consequently, the denoiser tends to over-compensate for a suboptimal predictor, leading to inefficient convergence. To address this, our DDC loss introduces explicit feature-level supervision. As reported in Tab.~\ref{tab:ablation-study}(g) vs (h), integrating $\mathcal{L}_{ddc}$ significantly boosts all quantitative metrics.
	Notably, an unconstrained predictor (g) even underperforms the baseline (f), proving DDC indispensable to unlock its capacity.
	
	This quantitative gain is intuitively corroborated by the internal representations shown in Fig.~\ref{fig:ablation-ddc}. Without DDC loss, the unconstrained predicted latent $x_p^{\tau_s}$ exhibits noisy outliers. When spatially visualized via VAE decoding, these outliers manifest as severe structural artifacts, ultimately causing a blurry final output. By explicitly aligning $x_p^{\tau_s}$ with the ground-truth manifold, our DDC loss regulates the latent space to prevent denoiser over-compensation, yielding highly realistic and rich textures.

\begin{table}[t]
\caption{Computational complexity comparisons. The runtimes are obtained on NVIDIA H20 GPU with videos of $1024^2$ pixels.}
\vspace{-2mm}
\renewcommand\arraystretch{1.1}
\center
\footnotesize
	\begin{tabular}{l|ccccc}
	\toprule
	 Methods & \makecell{Vivid-VR\\ \textit{(50 steps)}} & \makecell{DOVE\\ \textit{(1 step)}} & \makecell{Ours\\ \textit{(1 step)}} & \makecell{Ours\\ \textit{(5 steps)}}  \\ \hline
	Runtime per frame (s)    &  22.29   &  0.876   &  1.106   &  3.247   \\
	MANIQA $\uparrow$  &  0.376   &  0.320   &  0.402   &  \textbf{0.430}   \\
	MUSIQ $\uparrow$  &  67.61   &  57.82   &  68.52   &  \textbf{71.55}   \\
	CLIP-IQA $\uparrow$  &  0.450   &  0.353   &  0.571   &  \textbf{0.661}   \\
	DOVER $\uparrow$  &  \textbf{14.46}   &  11.84   &  13.40   &  14.37   \\
	\bottomrule
\end{tabular}
\label{tab:run-time}
\end{table}

	\vspace{-1mm}
	\subsection{Limitations}
	\label{sec:limitations}
	\vspace{-1mm}

	Although SATB-VR significantly accelerates the inference process compared to standard $50$-step diffusion paradigms, the computational latency for high-resolution videos remains a practical challenge. As shown in Tab.~\ref{tab:run-time}, our method exhibits a runtime disadvantage compared to aggressive single-step approaches. This is primarily attributed to the extra forward pass required by the auxiliary predictor, as well as the inherent footprint of the 5B-parameter DiT backbone and VAE decoding. However, we argue that this modest computational trade-off is highly justified, given the substantial gains in temporal consistency and fine-detailed quality it unlocks over $1$-step methods. Future work will explore lightweight architectures and VAE-free diffusion paradigms to further bridge this efficiency gap and facilitate real-time applications.

	\vspace{-2mm}
	\section{Conclusions}
	\label{sec:conclusions}
	\vspace{-2mm}
	
	We have proposed SATB-VR, a robust joint-optimization diffusion framework for the jump-starting video restoration. To resolve the train-inference discrepancy, we introduced the SATB training strategy, enabling effective co-optimization of an auxiliary predictor and an explicitly LQ-conditioned denoiser. Furthermore, we designed the DDC loss, which leverages the dynamically updated denoiser as an evaluator to explicitly align internal features, thereby significantly enhancing the predictor's estimation accuracy. Extensive quantitative and qualitative evaluations demonstrate the effectiveness of SATB-VR under few-step inference regimes (\eg, $\le 5$ steps).

	\clearpage
	{\small
		\bibliographystyle{ieee_fullname}
		\bibliography{egbib}

@String(IJCV = {Int. J. Comput. Vis.})

@String(CVPR= {IEEE Conf. Comput. Vis. Pattern Recog.})

@String(ICCV= {Int. Conf. Comput. Vis.})

@String(ECCV= {Eur. Conf. Comput. Vis.})

@String(ICLR = {Int. Conf. Learn. Represent.})

@String(AAAI = {AAAI})

@String(CVPRW= {IEEE Conf. Comput. Vis. Pattern Recog. Worksh.})

@String(CVPR  = {CVPR})

@String(ICCV  = {ICCV})

@String(ECCV  = {ECCV})

@String(ICLR  = {ICLR})

@String(CVPRW= {CVPRW})

@inproceedings{wang2019edvr,
  title={Edvr: Video restoration with enhanced deformable convolutional networks},
  author={Wang, Xintao and Chan, Kelvin CK and Yu, Ke and Dong, Chao and Change Loy, Chen},
  booktitle=CVPRW,
  pages={0--8},
  year={2019}
}

@inproceedings{chan2022investigating,
  title={Investigating tradeoffs in real-world video super-resolution},
  author={Chan, Kelvin CK and Zhou, Shangchen and Xu, Xiangyu and Loy, Chen Change},
  booktitle=CVPR,
  pages={5962--5971},
  year={2022}
}

@inproceedings{wang2021real,
  title={Real-esrgan: Training real-world blind super-resolution with pure synthetic data},
  author={Wang, Xintao and Xie, Liangbin and Dong, Chao and Shan, Ying},
  booktitle=CVPR,
  pages={1905--1914},
  year={2021}
}

@inproceedings{rombach2022high,
  title={High-resolution image synthesis with latent diffusion models},
  author={Rombach, Robin and Blattmann, Andreas and Lorenz, Dominik and Esser, Patrick and Ommer, Bj{\"o}rn},
  booktitle=CVPR,
  pages={10684--10695},
  year={2022}
}

@article{podell2023sdxl,
  title={Sdxl: Improving latent diffusion models for high-resolution image synthesis},
  author={Podell, Dustin and English, Zion and Lacey, Kyle and Blattmann, Andreas and Dockhorn, Tim and M{\"u}ller, Jonas and Penna, Joe and Rombach, Robin},
  journal={arXiv preprint arXiv:2307.01952},
  year={2023}
}

@article{blattmann2023stable,
  title={Stable video diffusion: Scaling latent video diffusion models to large datasets},
  author={Blattmann, Andreas and Dockhorn, Tim and Kulal, Sumith and Mendelevitch, Daniel and Kilian, Maciej and Lorenz, Dominik and Levi, Yam and English, Zion and Voleti, Vikram and Letts, Adam and others},
  journal={arXiv preprint arXiv:2311.15127},
  year={2023}
}

@article{wang2024exploiting,
  title={Exploiting diffusion prior for real-world image super-resolution},
  author={Wang, Jianyi and Yue, Zongsheng and Zhou, Shangchen and Chan, Kelvin CK and Loy, Chen Change},
  journal=IJCV,
  volume={132},
  number={12},
  pages={5929--5949},
  year={2024}
}

@inproceedings{yu2024scaling,
  title={Scaling up to excellence: Practicing model scaling for photo-realistic image restoration in the wild},
  author={Yu, Fanghua and Gu, Jinjin and Li, Zheyuan and Hu, Jinfan and Kong, Xiangtao and Wang, Xintao and He, Jingwen and Qiao, Yu and Dong, Chao},
  booktitle=CVPR,
  pages={25669--25680},
  year={2024}
}

@inproceedings{chen2025faithdiff,
  title={Faithdiff: Unleashing diffusion priors for faithful image super-resolution},
  author={Chen, Junyang and Pan, Jinshan and Dong, Jiangxin},
  booktitle=CVPR,
  pages={28188--28197},
  year={2025}
}

@inproceedings{zhou2024upscale,
  title={Upscale-a-video: Temporal-consistent diffusion model for real-world video super-resolution},
  author={Zhou, Shangchen and Yang, Peiqing and Wang, Jianyi and Luo, Yihang and Loy, Chen Change},
  booktitle=CVPR,
  pages={2535--2545},
  year={2024}
}

@inproceedings{yang2024motion,
  title={Motion-guided latent diffusion for temporally consistent real-world video super-resolution},
  author={Yang, Xi and He, Chenhang and Ma, Jianqi and Zhang, Lei},
  booktitle=ECCV,
  pages={224--242},
  year={2024},
  organization={Springer}
}

@inproceedings{peebles2023scalable,
  title={Scalable diffusion models with transformers},
  author={Peebles, William and Xie, Saining},
  booktitle=CVPR,
  pages={4195--4205},
  year={2023}
}

@article{yang2024cogvideox,
  title={Cogvideox: Text-to-video diffusion models with an expert transformer},
  author={Yang, Zhuoyi and Teng, Jiayan and Zheng, Wendi and Ding, Ming and Huang, Shiyu and Xu, Jiazheng and Yang, Yuanming and Hong, Wenyi and Zhang, Xiaohan and Feng, Guanyu and others},
  journal={arXiv preprint arXiv:2408.06072},
  year={2024}
}

@inproceedings{wang2025seedvr,
  title={Seedvr: Seeding infinity in diffusion transformer towards generic video restoration},
  author={Wang, Jianyi and Lin, Zhijie and Wei, Meng and Zhao, Yang and Yang, Ceyuan and Loy, Chen Change and Jiang, Lu},
  booktitle=CVPR,
  pages={2161--2172},
  year={2025}
}

@article{xie2025star,
  title={Star: Spatial-temporal augmentation with text-to-video models for real-world video super-resolution},
  author={Xie, Rui and Liu, Yinhong and Zhou, Penghao and Zhao, Chen and Zhou, Jun and Zhang, Kai and Zhang, Zhenyu and Yang, Jian and Yang, Zhenheng and Tai, Ying},
  journal={arXiv preprint arXiv:2501.02976},
  year={2025}
}

@article{wang2025seedvr2,
  title={Seedvr2: One-step video restoration via diffusion adversarial post-training},
  author={Wang, Jianyi and Lin, Shanchuan and Lin, Zhijie and Ren, Yuxi and Wei, Meng and Yue, Zongsheng and Zhou, Shangchen and Chen, Hao and Zhao, Yang and Yang, Ceyuan and others},
  journal={arXiv preprint arXiv:2506.05301},
  year={2025}
}

@article{chen2025dove,
  title={DOVE: Efficient One-Step Diffusion Model for Real-World Video Super-Resolution},
  author={Chen, Zheng and Zou, Zichen and Zhang, Kewei and Su, Xiongfei and Yuan, Xin and Guo, Yong and Zhang, Yulun},
  journal={arXiv preprint arXiv:2505.16239},
  year={2025}
}

@inproceedings{wu2023exploring,
  title={Exploring video quality assessment on user generated contents from aesthetic and technical perspectives},
  author={Wu, Haoning and Zhang, Erli and Liao, Liang and Chen, Chaofeng and Hou, Jingwen and Wang, Annan and Sun, Wenxiu and Yan, Qiong and Lin, Weisi},
  booktitle=ICCV,
  pages={20144--20154},
  year={2023}
}

@inproceedings{zhang2023md,
  title={MD-VQA: Multi-dimensional quality assessment for UGC live videos},
  author={Zhang, Zicheng and Wu, Wei and Sun, Wei and Tu, Danyang and Lu, Wei and Min, Xiongkuo and Chen, Ying and Zhai, Guangtao},
  booktitle=CVPR,
  pages={1746--1755},
  year={2023}
}

@article{loshchilov2017decoupled,
  title={Decoupled weight decay regularization},
  author={Loshchilov, Ilya and Hutter, Frank},
  journal={arXiv preprint arXiv:1711.05101},
  year={2017}
}

@article{lu2025dpm,
  title={Dpm-solver++: Fast solver for guided sampling of diffusion probabilistic models},
  author={Lu, Cheng and Zhou, Yuhao and Bao, Fan and Chen, Jianfei and Li, Chongxuan and Zhu, Jun},
  journal={arXiv preprint arXiv:2211.01095},
  year={2022}
}

@inproceedings{tao2017detail,
  title={Detail-revealing deep video super-resolution},
  author={Tao, Xin and Gao, Hongyun and Liao, Renjie and Wang, Jue and Jia, Jiaya},
  booktitle=ICCV,
  pages={4472--4480},
  year={2017}
}

@inproceedings{yi2019progressive,
  title={Progressive fusion video super-resolution network via exploiting non-local spatio-temporal correlations},
  author={Yi, Peng and Wang, Zhongyuan and Jiang, Kui and Jiang, Junjun and Ma, Jiayi},
  booktitle=ICCV,
  pages={3106--3115},
  year={2019}
}

@inproceedings{ke2021musiq,
  title={Musiq: Multi-scale image quality transformer},
  author={Ke, Junjie and Wang, Qifei and Wang, Yilin and Milanfar, Peyman and Yang, Feng},
  booktitle=ICCV,
  pages={5148--5157},
  year={2021}
}

@inproceedings{wang2023exploring,
  title={Exploring clip for assessing the look and feel of images},
  author={Wang, Jianyi and Chan, Kelvin CK and Loy, Chen Change},
  booktitle=AAAI,
  pages={2555--2563},
  year={2023}
}

@inproceedings{zhang2018unreasonable,
  title={The unreasonable effectiveness of deep features as a perceptual metric},
  author={Zhang, Richard and Isola, Phillip and Efros, Alexei A and Shechtman, Eli and Wang, Oliver},
  booktitle=CVPR,
  pages={586--595},
  year={2018}
}

@inproceedings{blau2018perception,
	title={The perception-distortion tradeoff},
	author={Blau, Yochai and Michaeli, Tomer},
	booktitle=CVPR,
	pages={6228--6237},
	year={2018}
}

@article{zhuang2025flashvsr,
  title={Flashvsr: Towards real-time diffusion-based streaming video super-resolution},
  author={Zhuang, Junhao and Guo, Shi and Cai, Xin and Li, Xiaohui and Liu, Yihao and Yuan, Chun and Xue, Tianfan},
  journal={arXiv preprint arXiv:2510.12747},
  year={2025}
}

@article{lv2026duo,
  title={DUO-VSR: Dual-Stream Distillation for One-Step Video Super-Resolution},
  author={Lv, Zhengyao and Xia, Menghan and Wang, Xintao and Wong, Kwan-Yee K},
  journal={arXiv preprint arXiv:2603.22271},
  year={2026}
}

@inproceedings{yue2025arbitrary,
  title={Arbitrary-steps image super-resolution via diffusion inversion},
  author={Yue, Zongsheng and Liao, Kang and Loy, Chen Change},
  booktitle=CVPR,
  pages={23153--23163},
  year={2025}
}

@article{cui2024taming,
  title={Taming diffusion prior for image super-resolution with domain shift sdes},
  author={Cui, Qinpeng and Liu, Yixuan and Zhang, Xinyi and Bao, Qiqi and Liao, Qingmin and Wang, Li and Lu, Tian and Liu, Zicheng and Wang, Zhongdao and Barsoum, Emad},
  journal={arXiv preprint arXiv:2409.17778},
  year={2024}
}

@inproceedings{hu2022lora,
  title={Lo{RA}: Low-Rank Adaptation of Large Language Models},
  author={Edward J Hu and Yelong Shen and Phillip Wallis and Zeyuan Allen-Zhu and Yuanzhi Li and Shean Wang and Lu Wang and Weizhu Chen},
  booktitle=ICLR,
  pages={12513--12525},
  year={2022}
}

@inproceedings{bai2025vivid,
  title={Vivid-VR: Distilling Concepts from Text-to-Video Diffusion Transformer for Photorealistic Video Restoration},
  author={Bai, Haoran and Chen, Xiaoxu and Yang, Canqian and He, Zongyao and Deng, Sibin and Chen, Ying},
  booktitle=ICLR,
  pages={1--16},
  year={2026}
}

@inproceedings{nan2025openvid,
  title={Openvid-1m: A large-scale high-quality dataset for text-to-video generation},
  author={Nan, Kepan and Xie, Rui and Zhou, Penghao and Fan, Tiehan and Yang, Zhenheng and Chen, Zhijie and Li, Xiang and Yang, Jian and Tai, Ying},
  booktitle=ICLR,
  pages={1045--1064},
  year={2025}
}

@inproceedings{chen2024sharegpt4video,
  title={Sharegpt4video: Improving video understanding and generation with better captions},
  author={Chen, Lin and Wei, Xilin and Li, Jinsong and Dong, Xiaoyi and Zhang, Pan and Zang, Yuhang and Chen, Zehui and Duan, Haodong and Lin, Bin and Tang, Zhenyu and others},
  booktitle=NeurIPS,
  pages={19472--19495},
  year={2024}
}

@inproceedings{wang2024internvid,
  title={Internvid: A large-scale video-text dataset for multimodal understanding and generation},
  author={Wang, Yi and He, Yinan and Li, Yizhuo and Li, Kunchang and Yu, Jiashuo and Ma, Xin and Li, Xinhao and Chen, Guo and Chen, Xinyuan and Wang, Yaohui and others},
  booktitle=ICLR,
  pages={42055--42079},
  year={2024}
}

@inproceedings{yang2022maniqa,
  title={Maniqa: Multi-dimension attention network for no-reference image quality assessment},
  author={Yang, Sidi and Wu, Tianhe and Shi, Shuwei and Lao, Shanshan and Gong, Yuan and Cao, Mingdeng and Wang, Jiahao and Yang, Yujiu},
  booktitle=CVPR,
  pages={1191--1200},
  year={2022}
}
	}

	\clearpage
	
	\appendix
	\section{Appendix}
	\label{sec:appendix}
	
	\vspace{-1mm}
	\subsection{Posterior-Inspired Derivation of SATB}
	\label{sec:appendix-derivation-satb}
	\vspace{-1mm}

    In this section, we present a posterior-inspired analysis to motivate the SNR-Aware Trajectory Blending (SATB) strategy. Specifically, we investigate how the predictor-induced deviation at the jump-start timestep propagates through the closed-form DDPM posterior mean. This formulation establishes the rationale for the SATB blending anchor and demonstrates that the constructed auxiliary trajectory maintains the standard diffusion variance schedule.
    
    It should be clarified that this derivation examines a constructed auxiliary trajectory rather than the native DDPM posterior. Since the jump-start state is inherently centered at the predictor output $x_p^{\tau_s}$ instead of being sampled directly from the forward process of $x_{gt}$, the standard posterior formulation does not strictly apply. Nevertheless, the closed-form posterior mean serves as a valuable analytical proxy, offering a grounded perspective on how the initial prediction error can naturally attenuate across denoising steps in a schedule-consistent manner.

    \vspace{1mm}
    \noindent\textbf{Preliminaries: DDPM Posterior Mean and Variance.}
    In the standard DDPM framework, the tractable posterior distribution is defined as:
    \begin{equation}
    	q(x_{t-1}\mid x_t, x_0)
    	=\mathcal{N}(x_{t-1}; \tilde{\mu}_t(x_t,x_0), \tilde{\beta}_t I),
    	\label{eq:appendix_tractable_posterior_distribution}
    \end{equation}
    where the posterior mean $\tilde{\mu}_t$ and variance $\tilde{\beta}_t$, conditioned on the initial state $x_0$ and the current noisy state $x_t$, are given by:
    \begin{equation}
    	\tilde{\mu}_t(x_t, x_0)
    	=\frac{\sqrt{\bar{\alpha}_{t-1}}\beta_t}{1-\bar{\alpha}_t}x_0+\frac{\sqrt{\alpha_t}(1-\bar{\alpha}_{t-1})}{1-\bar{\alpha}_t}x_t,
    	\label{eq:appendix_posterior_mean}
    \end{equation}
    \begin{equation}
    	\tilde{\beta}_t
    	=\frac{1-\bar{\alpha}_{t-1}}{1-\bar{\alpha}_t}\beta_t
    	=\frac{1-\bar{\alpha}_{t-1}}{1-\bar{\alpha}_t}(1-\alpha_t).
    	\label{eq:appendix_posterior_var}
    \end{equation}
    Here, $\beta_t$ denotes the predefined noise schedule at timestep $t$, with $\alpha_t = 1-\beta_t$ and $\bar{\alpha}_t = \prod_{i=1}^{t}\alpha_i$. Accordingly, the signal-to-noise ratio (SNR) at timestep $t$ is formulated as:
    \begin{equation}
    	\gamma_t = \frac{\bar{\alpha}_t}{1-\bar{\alpha}_t}.
    	\label{eq:appendix_signal-to-noise_ratio}
    \end{equation}

    \vspace{1mm}
    \noindent\textbf{Constructed Jump-Start State.}
    To simulate the jump-starting inference paradigm during training, we formulate the initial noisy latent at timestep $\tau_s$ by anchoring the diffusion noise to the predictor output $x_p^{\tau_s}$:
    \begin{equation}
    	x_{\tau_s}
    	=\sqrt{\bar{\alpha}_{\tau_s}}\,x_p^{\tau_s}+\sqrt{1-\bar{\alpha}_{\tau_s}}\,\epsilon,\quad \epsilon \sim \mathcal{N}(0,I).
    	\label{eq:appendix_x_taus}
    \end{equation}
    By defining the predictor-induced deviation as:
    \begin{equation}
    	\delta_{\tau_s} = x_p^{\tau_s} - x_{gt}.
    	\label{eq:appendix_delta}
    \end{equation}
    Eq.~\eqref{eq:appendix_x_taus} can be equivalently reparameterized as:
    \begin{equation}
    	x_{\tau_s}
    	=\sqrt{\bar{\alpha}_{\tau_s}}(x_{gt}+\delta_{\tau_s})+\sqrt{1-\bar{\alpha}_{\tau_s}}\,\epsilon,\quad \epsilon \sim \mathcal{N}(0,I).
    \end{equation}
    Although this constructed latent diverges from the standard forward marginal $q(x_{\tau_s}\mid x_{gt})$, substituting it into the tractable DDPM posterior mean reveals how the predictor-induced deviation propagates during the denoising step.

    \vspace{1mm}
    \noindent\textbf{One-Step Analysis of Predictor-Deviation Attenuation.}
    During joint optimization, the denoiser must robustly compensate for initial prediction deviation while smoothly transitioning the trajectory from the predictor output $x_p^{\tau_s}$ toward the true data manifold $x_{gt}$. By anchoring the clean endpoint at $x_{gt}$ and substituting Eq.~\eqref{eq:appendix_x_taus} into Eq.~\eqref{eq:appendix_posterior_mean}, we expand the posterior mean as:
    \begin{align}
    	\tilde{\mu}_{\tau_s}&(x_{\tau_s}, x_{gt}) \notag \\
    	&= \frac{\sqrt{\bar{\alpha}_{{\tau_s}-1}}\beta_{\tau_s}}{1-\bar{\alpha}_{\tau_s}}x_{gt} \notag \\
    	&\quad + \frac{\sqrt{\alpha_{\tau_s}}(1-\bar{\alpha}_{{\tau_s}-1})}{1-\bar{\alpha}_{\tau_s}} \left( \sqrt{\bar{\alpha}_{\tau_s}} x_p^{\tau_s} + \sqrt{1-\bar{\alpha}_{\tau_s}}\, \epsilon \right) \notag \\
    	&= \frac{\sqrt{\bar{\alpha}_{{\tau_s}-1}}(1-\alpha_{\tau_s})}{1-\bar{\alpha}_{\tau_s}}x_{gt} \notag  + \frac{(1-\bar{\alpha}_{{\tau_s}-1})}{1-\bar{\alpha}_{\tau_s}}\sqrt{\alpha_{\tau_s}\bar{\alpha}_{\tau_s}}\, x_p^{\tau_s} \notag \\
    	&\quad + \frac{\sqrt{\alpha_{\tau_s}}(1-\bar{\alpha}_{{\tau_s}-1})}{1-\bar{\alpha}_{\tau_s}}\sqrt{1-\bar{\alpha}_{\tau_s}}\, \epsilon \notag \\
    	&= \sqrt{\bar{\alpha}_{\tau_s-1}}\left[\,\frac{1-\alpha_{\tau_s}}{1-\bar{\alpha}_{\tau_s}}\,x_{gt} + \frac{\alpha_{\tau_s}(1-\bar{\alpha}_{{\tau_s}-1})}{(1-\bar{\alpha}_{\tau_s})}\, x_p^{\tau_s}\right] \notag \\
    	&\quad + \frac{\sqrt{\alpha_{\tau_s}}(1-\bar{\alpha}_{{\tau_s}-1})}{1-\bar{\alpha}_{\tau_s}}\sqrt{1-\bar{\alpha}_{\tau_s}}\, \epsilon.
    	\label{eq:appendix_expanded_mean}
    \end{align}
    Grouping the noise-free terms ($x_{gt}$ and $x_p^{\tau_s}$) and factoring out the common scale $\sqrt{\bar{\alpha}_{\tau_s-1}}$, the coefficient for $x_{gt}$ simplifies to:
    \begin{align}
    	\frac{1-\alpha_{\tau_s}}{1-\bar{\alpha}_{\tau_s}}
    	&=\frac{\bar{\alpha}_{\tau_s-1}(1-\alpha_{\tau_s})}{\bar{\alpha}_{\tau_s-1}(1-\bar{\alpha}_{\tau_s})}\notag\\
    	&=\frac{\bar{\alpha}_{\tau_s-1}-\bar{\alpha}_{\tau_s}}{\bar{\alpha}_{\tau_s-1}(1-\bar{\alpha}_{\tau_s})}
    	\notag\\
    	&=\frac{\bar{\alpha}_{\tau_s-1}(1-\bar{\alpha}_{\tau_s})-\bar{\alpha}_{\tau_s}(1-\bar{\alpha}_{\tau_s-1})}
    	{\bar{\alpha}_{\tau_s-1}(1-\bar{\alpha}_{\tau_s})}
    	\notag\\
    	&=1-\frac{\bar{\alpha}_{\tau_s}(1-\bar{\alpha}_{\tau_s-1})}
    	{\bar{\alpha}_{\tau_s-1}(1-\bar{\alpha}_{\tau_s})}\notag\\
    	&=1-\frac{\gamma_{\tau_s}}{\gamma_{\tau_s-1}}.
    	\label{eq:appendix_xgt_coeff}
    \end{align}
    Similarly, the coefficient for $x_p^{\tau_s}$ exactly reduces to the relative SNR:
    \begin{align}
    	\frac{\alpha_{\tau_s}(1-\bar{\alpha}_{\tau_s-1})}{1-\bar{\alpha}_{\tau_s}}
    	&=\frac{\bar{\alpha}_{\tau_s}(1-\bar{\alpha}_{\tau_s-1})}
    	{\bar{\alpha}_{\tau_s-1}(1-\bar{\alpha}_{\tau_s})}\notag\\
    	&=\frac{\bar{\alpha}_{\tau_s}/(1-\bar{\alpha}_{\tau_s})}
    	{\bar{\alpha}_{\tau_s-1}/(1-\bar{\alpha}_{\tau_s-1})}\notag\\
    	&=\frac{\gamma_{\tau_s}}{\gamma_{\tau_s-1}}.
    	\label{eq:appendix_xp_coeff}
    \end{align}
    Substituting Eq.~\eqref{eq:appendix_xgt_coeff} and Eq.~\eqref{eq:appendix_xp_coeff} back into Eq.~\eqref{eq:appendix_expanded_mean} yields:
    \begin{align}
    	\tilde{\mu}_{\tau_s}(x_{\tau_s}, x_{gt})
    	&=\sqrt{\bar{\alpha}_{\tau_s-1}}\left[\left(1-\frac{\gamma_{\tau_s}}{\gamma_{\tau_s-1}}\right)x_{gt}+\frac{\gamma_{\tau_s}}{\gamma_{\tau_s-1}}x_p^{\tau_s}\right]\notag\\
    	&\quad+\frac{\sqrt{\alpha_{\tau_s}}(1-\bar{\alpha}_{{\tau_s}-1})}{1-\bar{\alpha}_{\tau_s}}\sqrt{1-\bar{\alpha}_{\tau_s}}\,\epsilon \notag\\
    	&=\sqrt{\bar{\alpha}_{\tau_s-1}}\left[x_{gt}+\frac{\gamma_{\tau_s}}{\gamma_{\tau_s-1}}(x_p^{\tau_s}-x_{gt})\right]\notag\\
    	&\quad + \frac{\sqrt{\alpha_{\tau_s}}(1-\bar{\alpha}_{{\tau_s}-1})}{1-\bar{\alpha}_{\tau_s}} \sqrt{1-\bar{\alpha}_{\tau_s}}\,\epsilon.
    	\label{eq:appendix_mean_final}
    \end{align}
    By matching the structure of Eq.~\eqref{eq:appendix_mean_final} with Eq.~\eqref{eq:appendix_x_taus}, the bracketed term is directly identified as the corresponding auxiliary anchor at timestep $\tau_s-1$:
    \begin{equation}
    	x_b^{\tau_s-1}
    	=x_{gt}+\frac{\gamma_{\tau_s}}{\gamma_{\tau_s-1}}(x_p^{\tau_s}-x_{gt}).
    	\label{eq:appendix_blended_anchor_one_step}
    \end{equation}
    Expressed in terms of $\delta_{\tau_s}$ defined in Eq.~\eqref{eq:appendix_delta}, this becomes:
    \begin{equation}
    	x_b^{\tau_s-1}
    	=x_{gt}+\frac{\gamma_{\tau_s}}{\gamma_{\tau_s-1}}\delta_{\tau_s}.
    	\label{eq:appendix_blended_anchor_one_step_delta}
    \end{equation}
    This reveals that, during a single reverse step from $\tau_s$ to $\tau_s-1$, the initial prediction deviation from the true manifold is naturally attenuated by the relative SNR.

    \vspace{1mm}
    \noindent\textbf{Variance-Schedule Consistency.}
    Crucially, the constructed auxiliary trajectory must strictly preserve the predefined noise schedule. To verify this consistency for the one-step transition above, we evaluate the total variance $\hat{\beta}_{\tau_s}$ of the latent at step $\tau_s-1$. Given the additivity of independent Gaussian components, $\hat{\beta}_{\tau_s}$ is the sum of the residual variance from Eq.~\eqref{eq:appendix_mean_final} and the posterior variance $\tilde{\beta}_{\tau_s}$ from Eq.~\eqref{eq:appendix_posterior_var}:
    \begin{align}
    	\hat{\beta}_{\tau_s}
    	&=\left(\frac{\sqrt{\alpha_{\tau_s}}(1-\bar{\alpha}_{{\tau_s}-1})}{1-\bar{\alpha}_{\tau_s}}\sqrt{1-\bar{\alpha}_{\tau_s}}\right)^2+\tilde{\beta}_{\tau_s}\notag\\
    	&=\frac{\alpha_{\tau_s}(1-\bar{\alpha}_{{\tau_s}-1})^2}{1-\bar{\alpha}_{\tau_s}}+\frac{1-\bar{\alpha}_{{\tau_s}-1}}{1-\bar{\alpha}_{\tau_s}}(1-\alpha_{\tau_s})\notag\\
    	&=
    	\frac{1-\bar{\alpha}_{{\tau_s}-1}}{1-\bar{\alpha}_{\tau_s}}\left(\alpha_{\tau_s}(1-\bar{\alpha}_{{\tau_s}-1}) + 1-\alpha_{\tau_s}\right) \notag\\
    	&=\frac{1-\bar{\alpha}_{{\tau_s}-1}}{1-\bar{\alpha}_{\tau_s}}\left(1-\alpha_{\tau_s}\bar{\alpha}_{{\tau_s}-1}\right)\notag\\
    	&=1-\bar{\alpha}_{{\tau_s}-1}.
    	\label{eq:appendix_variance_proof}
    \end{align}
    This confirms that the constructed auxiliary latent retains the standard diffusion variance. Consequently, it can be reparameterized as:
    \begin{equation}
    	x_{\tau_s-1}
    	=
    	\sqrt{\bar{\alpha}_{\tau_s-1}}\,x_b^{\tau_s-1}
    	+
    	\sqrt{1-\bar{\alpha}_{\tau_s-1}}\,\epsilon,
    	\quad
    	\epsilon\sim\mathcal{N}(0,I),
    	\label{eq:appendix_q_taus_minus_1}
    \end{equation}
    inducing the corresponding auxiliary distribution:
    \begin{align}
    	\hat q&(x_{\tau_s-1}\mid x_p^{\tau_s}, x_{gt})\notag\\
    	&=
    	\mathcal{N}
    	\left(
    	x_{\tau_s-1};
    	\sqrt{\bar{\alpha}_{\tau_s-1}}\,x_b^{\tau_s-1},
    	(1-\bar{\alpha}_{\tau_s-1})I
    	\right).
    	\label{eq:appendix_q_taus_minus_1_dist}
    \end{align}

    \vspace{1mm}
    \noindent\textbf{Generalization to Arbitrary Timesteps.}
    Generalizing this one-step transition to an arbitrary timestep $t \le \tau_s$, the sequential SNR-ratio attenuation forms a telescoping product:
    \begin{equation}
    \prod_{i=t+1}^{\tau_s} \frac{\gamma_i}{\gamma_{i-1}} = \frac{\gamma_{\tau_s}}{\gamma_t}. 
    \end{equation}
    This directly yields the generalized auxiliary anchor, formally deriving the SATB formulation presented in Eq.~\eqref{eq:satb-blend} of the main paper:
    \begin{equation}
    	x_b^t
    	=x_{gt}+\frac{\gamma_{\tau_s}}{\gamma_t}(x_p^{\tau_s}-x_{gt}).
    	\label{eq:appendix_xbt}
    \end{equation}
    Notably, this formulation satisfies the exact boundary condition:
    \begin{equation}
    	x_b^{\tau_s} = x_p^{\tau_s},
    \end{equation}
    and progressively converges to $x_{gt}$ at earlier, higher-SNR timesteps. Assuming a standard monotonic noise schedule, the scaling coefficient strictly satisfies:
    \begin{equation}
    	0 \le \frac{\gamma_{\tau_s}}{\gamma_t} \le 1,
    \end{equation}
    confirming that $x_b^t$ forms a valid geometric contraction from the predictor estimate $x_p^{\tau_s}$ toward the true manifold $x_{gt}$.
    
    Consequently, the auxiliary latent distribution at any timestep $t$ is defined as:
    \begin{equation}
    	\hat q(x_t\mid x_p^{\tau_s}, x_{gt})
        =\mathcal{N}\left(x_t;\sqrt{\bar{\alpha}_t}\,x_b^t,(1-\bar{\alpha}_t)I\right),
    	\label{eq:appendix_generalization_dist}
    \end{equation}
    yielding the closed-form reparameterization:
    \begin{equation}
    	x_t = \sqrt{\bar{\alpha}_t}\left(x_{gt}+\frac{\gamma_{\tau_s}}{\gamma_t}(x_p^{\tau_s}-x_{gt})\right)+\sqrt{1-\bar{\alpha}_t}\,\epsilon.
    	\label{eq:appendix_generalization}
    \end{equation}
    Thus, SATB constructs an auxiliary training trajectory whose signal center gradually moves from the initial predictor output back to the ground-truth anchor, while preserving the native diffusion variance schedule.
    
    \vspace{1mm}
    \noindent\textbf{Conclusion.}
    The above analysis provides a mathematical foundation for the SATB strategy. Rather than an empirical heuristic, the relative SNR coefficient $\gamma_{\tau_s}/\gamma_t$ emerges naturally from the tractable posterior mean, characterizing the attenuation of initial prediction errors across denoising steps. Crucially, the constructed auxiliary trajectory strictly preserves the native diffusion variance $1-\bar{\alpha}_t$ at each timestep. Consequently, perturbing the SATB anchor (Eq.~\eqref{eq:satb-blend} of the main paper) with standard forward noise yields a schedule-consistent training input. This effectively compels the denoiser to rectify the initial predictor bias at $t=\tau_s$, while steering the trajectory back to the ground-truth data manifold as the SNR increases.

	\vspace{-1mm}
	\subsection{More Implementation Details}
	\label{sec:appendix-more-implementation-details}
	\vspace{-1mm}

	\vspace{1mm}
	\noindent\textbf{Training Data Preparation.}
	To facilitate robust training, we curate a large-scale, high-quality video-text dataset. We initially collect approximately $2.4M$ source videos, comprising $1M$ videos from OpenVid-1M~\cite{nan2025openvid}, $0.4M$ from ShareGPT4Video~\cite{chen2024sharegpt4video}, and the first $1M$ videos from InternVid~\cite{wang2024internvid}. 
	While these datasets have undergone preliminary curation by their respective authors, we observe residual scene transitions that may disrupt temporal consistency during training. Consequently, we employ PySceneDetect to perform scene transition detection and clip segmentation. We then filter the resulting clips based on physical constraints, retaining only those with a resolution greater than $1024$ (both width and height) and a duration exceeding $2$ seconds. 
	To guarantee visual quality, we further apply the no-reference metric MD-VQA~\cite{zhang2023md} as a quality filter, setting a strict threshold of $>90$. This pipeline yields a highly refined subset of $200K$ videos. For text conditioning, we re-caption these videos using CogVLM2-Video~\cite{yang2024cogvideox} with the system prompt: \textit{``Please describe this video in detail.''}
	
	During the joint training phase, we resize the short side of training videos to $1024$ pixels and then center-crop them to a resolution of $1024 \times 1024$. The number of frames for the training videos is randomly selected between $17$ and $37$.
	We adopt the degradation pipeline from Real-ESRGAN~\cite{wang2021real} to synthesize the corresponding low-quality video inputs. 
	Furthermore, to alleviate the distribution drift issue during base model fine-tuning, we incorporate $50K$ synthetic videos generated via the concept distillation strategy proposed in~\cite{bai2025vivid}, mixing them into our training pool.

	\vspace{1mm}
	\noindent\textbf{Inference Scheduling.}
	During inference, our SATB-VR framework supports a flexible few-step regime. We dynamically adjust the denoising starting point $\tau_s$ based on the desired number of total inference steps. Specifically, a larger total step budget allows for a higher $\tau_s$, which injects more noise and consequently unlocks stronger generative capacity for recovering high-frequency fine details. Once $\tau_s$ is determined, the subsequent inference steps are uniformly sampled across the remaining diffusion trajectory. The detailed timestep scheduling sequences for various inference steps ($1 \sim 10$ steps) are summarized in Tab.~\ref{tab:timestep-schedule}.

\begin{table}[h]
\caption{Detailed timestep scheduling for different inference steps.}
\vspace{-2mm}
\renewcommand\arraystretch{1.1}
\center
\footnotesize
\begin{tabular}{c c l}
	\toprule
	\# Steps & $\tau_s$ & Inference timestep schedule $\{\tau_i\}_{i=1}^s$ \\
	\midrule
	1  & 199 & \{199\} \\
	2  & 199 & \{199, 99\} \\
	3  & 299 & \{299, 199, 99\} \\
	4  & 399 & \{399, 299, 199, 99\} \\
	5  & 399 & \{399, 319, 239, 159, 79\} \\
	8  & 399 & \{399, 349, 299, 249, 199, 149, 99, 49\} \\
	10 & 399 & \{399, 359, 319, 279, 239, 199, 159, 119, 79, 39\} \\
	\bottomrule
\end{tabular}
\label{tab:timestep-schedule}
\end{table}

	\vspace{-1mm}
	\subsection{Effect of the Inference Steps}
	\label{sec:appendix-effect-inference-steps}
	\vspace{-1mm}
	
\begin{figure*}[!t]
		\centering
		\includegraphics[width=1.0\linewidth]{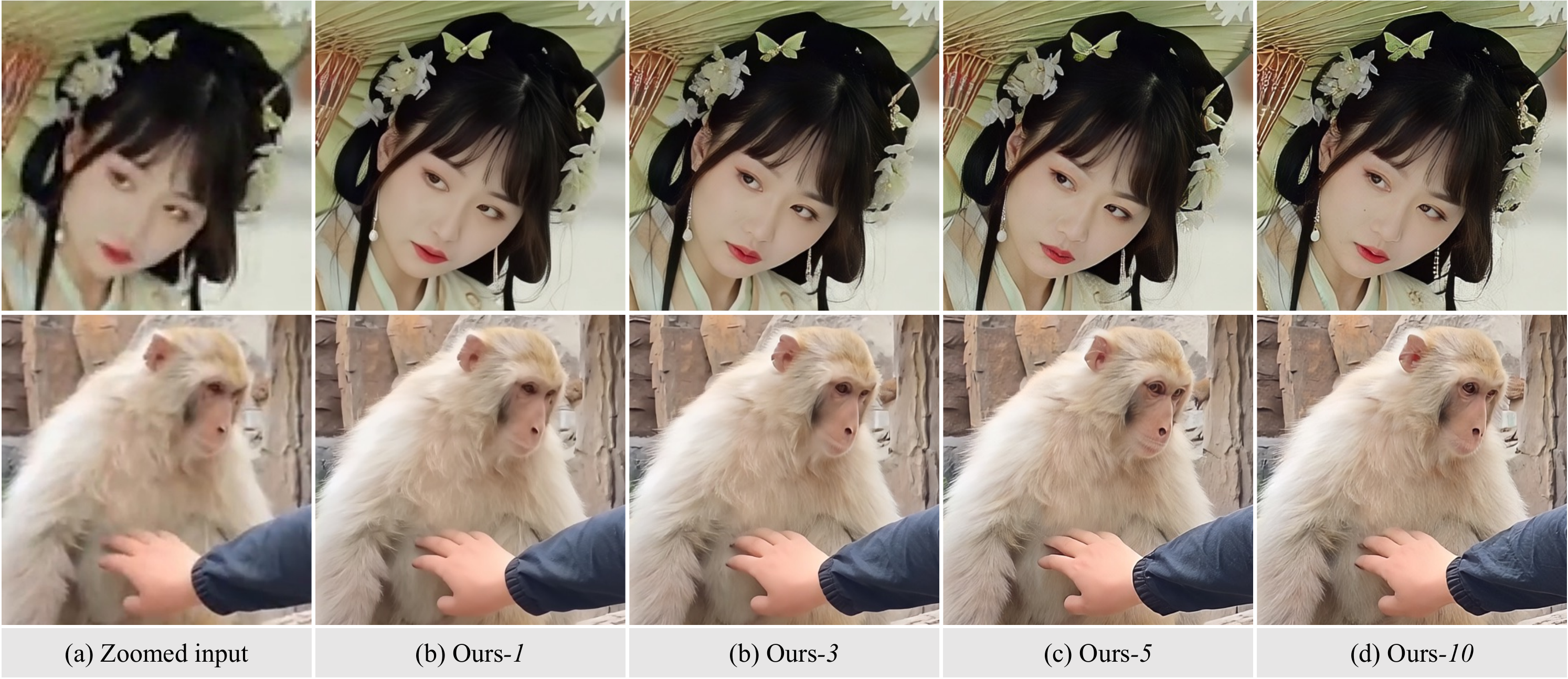}
		\caption{%
			Performance trends at various inference steps. Increasing the number of inference steps unlocks stronger generative capability, enabling the method to progressively produce more realistic fine textures. (\textbf{Zoom-in for best view})
		}
		\label{fig:supp-inference-steps}
	\end{figure*}

	In the main paper, we have shown the performance trends at various inference steps. Here, we further provide a comprehensive evaluation regarding the impact of varying inference steps. Tab.~\ref{tab:steps-quantitative-results} details the quantitative performance of our SATB-VR across all benchmarks under different inference steps (\eg, from $1$ to $10$ steps). 
	In Fig.~\ref{fig:supp-inference-steps}, we further provide more visualization results, where increasing the number of inference steps unlocks stronger generative capacity. Consequently, the proposed method progressively produces more realistic fine textures.

\begin{table*}[!t]
\caption{Quantitative evaluation of our SATB-VR across all benchmarks under different inference steps. The best and second performances are marked in {\color[HTML]{F94848} red} and {\color[HTML]{3166FF} blue}, respectively.}
\vspace{1mm}
\renewcommand\arraystretch{1.1}
\footnotesize
\begin{tabular}{c|c|cc|ccccccc}
\toprule
Datasets    & Metrics    & \makecell{~Vivid-VR~ \\ (\textit{50 steps})}    & \makecell{~~~DOVE~~~ \\ (\textit{1 step})}    & \makecell{~~~~Ours~~~~ \\ (\textit{1 step})}    & \makecell{~~~~Ours~~~~ \\ (\textit{2 steps})}     & \makecell{~~~~Ours~~~~ \\ (\textit{3 steps})}     &\makecell{~~~~Ours~~~~ \\ (\textit{4 steps})}     & \makecell{~~~~Ours~~~~ \\ (\textit{5 steps})}    & \makecell{~~~~Ours~~~~ \\ (\textit{8 steps})}     & \makecell{~~~~Ours~~~~ \\ (\textit{10 steps})}     \\ \hline
       & PSNR $\uparrow$ & 21.73 & {\color[HTML]{F94848} 24.80} & {\color[HTML]{3166FF} 24.18} & 23.81 & 23.17 & 22.06 & 21.69 & 21.64 & 21.55 \\ 
       & SSIM $\uparrow$ & 0.604 & {\color[HTML]{F94848} 0.754} & {\color[HTML]{3166FF} 0.707} & 0.691   & 0.664   & 0.612   & 0.599 & 0.589   & 0.587   \\ 
       & LPIPS $\downarrow$ & 0.278 & {\color[HTML]{F94848} 0.168} & {\color[HTML]{3166FF} 0.197} & 0.209   & 0.233   & 0.285   & 0.294 & 0.312   & 0.314   \\ \cline{2-11} 
       & MANIQA $\uparrow$ & 0.410 & 0.346 & 0.384 & 0.388   & 0.403   & {\color[HTML]{3166FF} 0.424}   & {\color[HTML]{F94848} 0.433} & 0.420   & 0.423   \\ 
       & MUSIQ $\uparrow$ & 70.03 & 63.29 & 67.82 & 68.71 & 70.52 & 71.57 & {\color[HTML]{F94848} 72.14} & 71.49 & {\color[HTML]{3166FF} 71.69} \\ 
       & CLIP-IQA $\uparrow$ & 0.483 & 0.410 & 0.514 & 0.548   & 0.581   & {\color[HTML]{3166FF} 0.623}   & {\color[HTML]{F94848} 0.625} & 0.605   & 0.609   \\ 
\multirow{-7}{*}{SPMCS}  & DOVER $\uparrow$    & 11.35 & 9.898 & 10.65 & 11.42 & 12.16 & 11.91 & 11.93 & {\color[HTML]{3166FF} 12.28} & {\color[HTML]{F94848} 12.34} \\ \hline
        & PSNR $\uparrow$ & 24.54 & {\color[HTML]{F94848} 30.53} & {\color[HTML]{3166FF} 28.67} & 28.05 & 27.12 & 25.90 & 25.66 & 25.57 & 25.47 \\ 
        & SSIM $\uparrow$ & 0.761 & {\color[HTML]{F94848} 0.894} & {\color[HTML]{3166FF} 0.859} & 0.843   & 0.819   & 0.780   & 0.772 & 0.765   & 0.762   \\ 
        & LPIPS $\downarrow$ & 0.243 & {\color[HTML]{F94848} 0.101} & {\color[HTML]{3166FF} 0.150} & 0.160   & 0.182   & 0.219   & 0.229 & 0.233   & 0.235   \\ \cline{2-11} 
        & MANIQA $\uparrow$ & 0.359 & 0.296 & 0.381 & 0.376   & 0.391   & {\color[HTML]{3166FF} 0.416}   & {\color[HTML]{F94848} 0.416} & 0.410   & 0.412   \\ 
        & MUSIQ $\uparrow$ & 64.71 & 55.17 & 65.83 & 66.15 & 67.74 & 69.16 & {\color[HTML]{F94848} 69.75} & 69.34 & {\color[HTML]{3166FF} 69.61} \\ 
        & CLIP-IQA $\uparrow$ & 0.426 & 0.340 & 0.507 & 0.510   & 0.544   & 0.597   & {\color[HTML]{3166FF} 0.601} & 0.601   & {\color[HTML]{F94848} 0.604}   \\ 
\multirow{-7}{*}{UDM10}       & DOVER $\uparrow$ & 11.97 & 10.41 & 10.98 & 11.52 & 11.82 & 12.44 & 12.49 & {\color[HTML]{3166FF} 12.68} & {\color[HTML]{F94848} 12.73} \\ \hline
        & PSNR $\uparrow$ & 21.31 & {\color[HTML]{F94848} 24.10} & {\color[HTML]{3166FF} 23.67} & 23.29 & 22.81 & 22.13 & 21.98 & 21.85 & 21.82 \\ 
        & SSIM $\uparrow$ & 0.579 & {\color[HTML]{F94848} 0.688} & {\color[HTML]{3166FF} 0.657} & 0.644   & 0.627   & 0.595   & 0.589 & 0.581   & 0.580   \\ 
        & LPIPS $\downarrow$ & 0.357 & 0.283 & {\color[HTML]{3166FF} 0.281} & {\color[HTML]{F94848} 0.280}   & 0.288   & 0.301   & 0.303 & 0.307   & 0.309   \\ \cline{2-11} 
        & MANIQA $\uparrow$ & 0.372 & 0.304 & 0.354  & 0.351   & 0.363   & 0.385   & {\color[HTML]{F94848} 0.396} & 0.382   & {\color[HTML]{3166FF} 0.388}   \\ 
        & MUSIQ $\uparrow$ & 70.55 & 60.65 & 67.91 & 68.94 & 70.50 & 71.72 & {\color[HTML]{3166FF} 72.34} & 72.09 & {\color[HTML]{F94848} 72.38} \\ 
        & CLIP-IQA $\uparrow$ & 0.447 & 0.356 & 0.486 & 0.517   & 0.550   & 0.597   & {\color[HTML]{F94848} 0.603} & 0.595   & {\color[HTML]{3166FF} 0.599}   \\ 
\multirow{-7}{*}{YouHQ40}     & DOVER $\uparrow$ & 14.61 & 12.52 & 13.25 & 13.85 & 14.21 & 14.19 & 14.46 & {\color[HTML]{3166FF} 14.76} & {\color[HTML]{F94848} 14.81} \\ \hline
        & MANIQA $\uparrow$ & 0.319 & 0.272 & 0.356 & 0.350   & 0.363   & 0.381   & {\color[HTML]{F94848} 0.383} & 0.379   & {\color[HTML]{3166FF} 0.381}   \\ 
        & MUSIQ $\uparrow$ & 62.47 & 55.11 & 65.59 & 65.88   & 67.01 & 68.03 & 68.28 & {\color[HTML]{3166FF} 68.31} & {\color[HTML]{F94848} 68.48} \\ 
        & CLIP-IQA $\uparrow$ & 0.338 & 0.271 & 0.436 & 0.446   & 0.461   & {\color[HTML]{F94848} 0.488}   & 0.485 & 0.484   & {\color[HTML]{3166FF} 0.486}   \\ 
\multirow{-4}{*}{VideoLQ}     & DOVER $\uparrow$ & 9.743 & 8.780 & 9.577 & 9.688   & 9.845   & 9.846   & 9.930 & {\color[HTML]{3166FF} 9.939}   & {\color[HTML]{F94848} 10.04} \\ \hline
        & MANIQA $\uparrow$ & 0.376 & 0.320 & 0.402 & 0.394   & 0.410   & {\color[HTML]{3166FF} 0.427}   & {\color[HTML]{F94848} 0.430} & 0.421   & 0.424   \\ 
        & MUSIQ $\uparrow$ & 67.61 & 57.82 & 68.52 & 68.95   & 70.43 & 71.12 & {\color[HTML]{3166FF} 71.55} & 71.41 & {\color[HTML]{F94848} 71.67} \\ 
        & CLIP-IQA $\uparrow$ & 0.450 & 0.353 & 0.571 & 0.593   & 0.624   & 0.659   & 0.661 & {\color[HTML]{3166FF} 0.661}   & {\color[HTML]{F94848} 0.663}   \\ 
\multirow{-4}{*}{UGC50}       & DOVER $\uparrow$ & 14.46 & 11.84 & 13.40 & 14.01   & 14.33 & 14.35 & 14.37 & {\color[HTML]{F94848} 14.77} & {\color[HTML]{3166FF} 14.75} \\ \hline
        & MANIQA $\uparrow$ & 0.369 & 0.334 & 0.378 & 0.376   & 0.391   & {\color[HTML]{3166FF} 0.415}   & {\color[HTML]{F94848} 0.415} & 0.411   & 0.412   \\ 
        & MUSIQ $\uparrow$ & 67.18 & 62.07 & 66.08 & 66.93   & 68.79 & 70.24 & {\color[HTML]{3166FF} 70.60} & 70.57 & {\color[HTML]{F94848} 70.79} \\ 
        & CLIP-IQA $\uparrow$ & 0.445 & 0.379 & 0.493 & 0.506   & 0.539   & 0.591   & 0.594 & {\color[HTML]{3166FF} 0.599}   & {\color[HTML]{F94848} 0.602}   \\ 
\multirow{-4}{*}{AIGC50}      & DOVER $\uparrow$ & 14.51 & 14.49 & 14.43 & 14.66  & 15.04 & 15.10 & 15.14 & {\color[HTML]{F94848} 15.37} & {\color[HTML]{3166FF} 15.23} \\
\bottomrule
\end{tabular}
\label{tab:steps-quantitative-results}
\end{table*}

	\vspace{-1mm}
	\subsection{More Visualization Results}
	\label{sec:appendix-more-visualization-results}
	\vspace{-1mm}
	
\begin{figure*}[!t]
		\centering
		\includegraphics[width=1.0\linewidth]{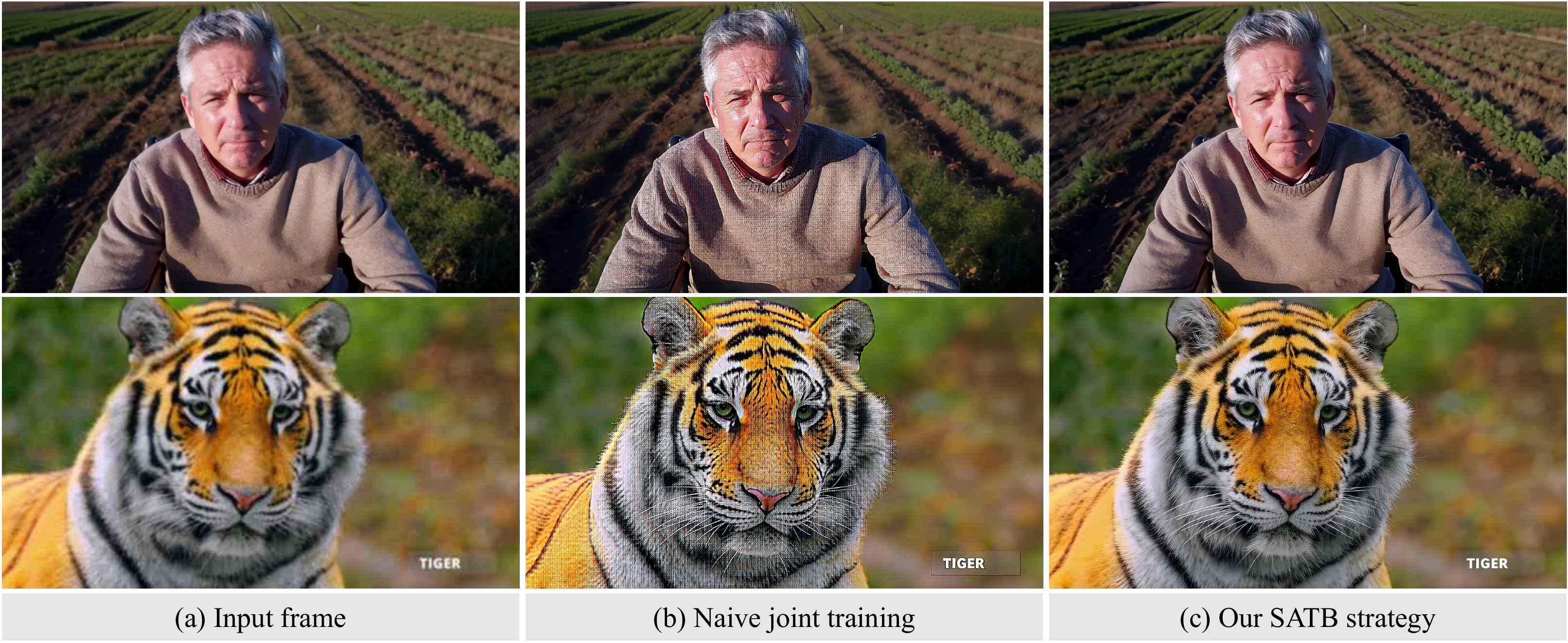}
		\caption{%
			Additional visual demonstration of the SATB strategy. By effectively bridging the train-inference gap, our proposed SATB strategy enables robust joint training and yields high-quality, artifact-free restoration results. (\textbf{Zoom-in for best view})
		}
		\label{fig:supp-visual-artifact}
	\end{figure*}

\begin{figure*}[!t]
		\centering
		\includegraphics[width=1.0\linewidth]{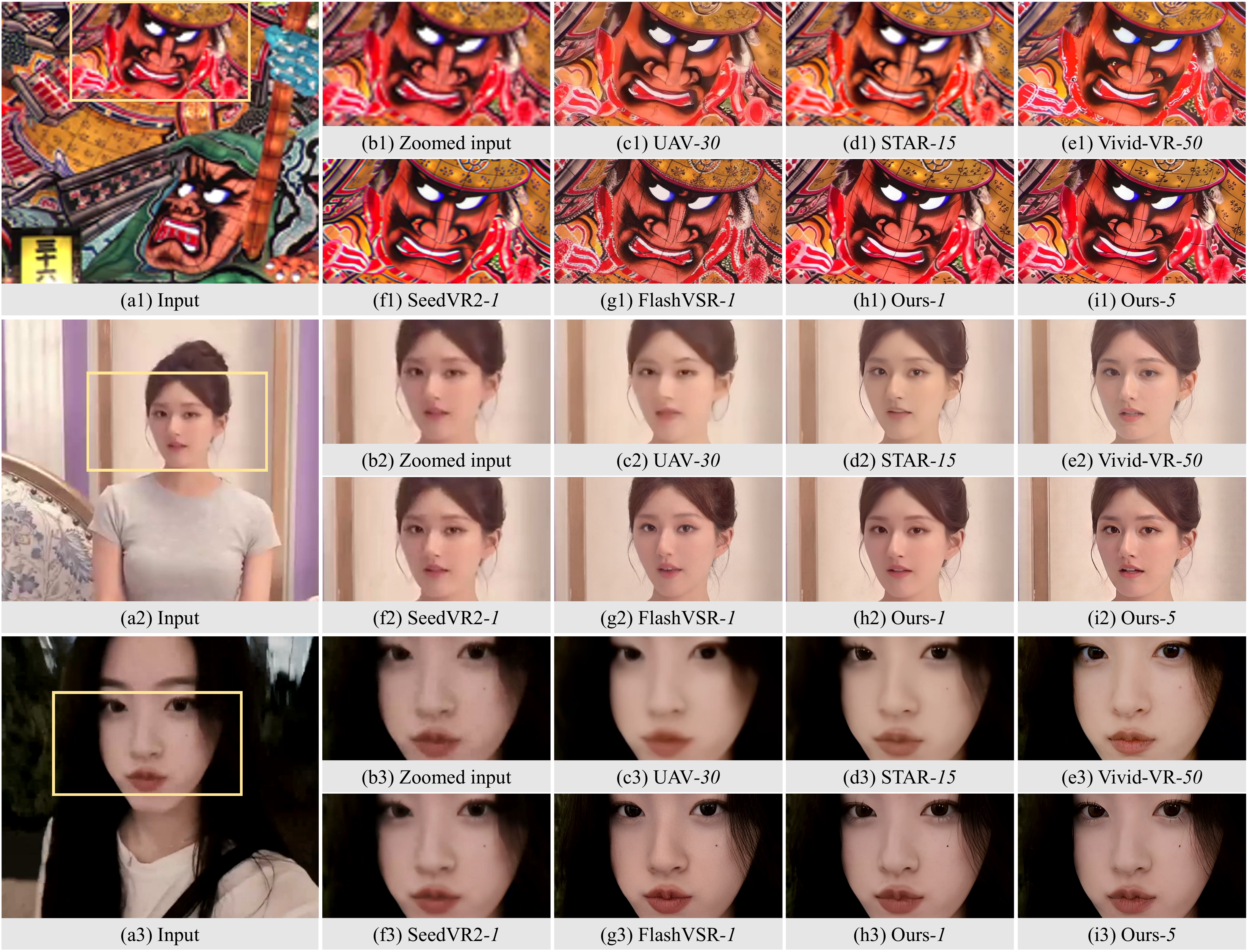}
		\caption{%
			Qualitative comparison results on synthetic ($1st$ row) and real-world ($2nd$ and $3rd$ rows) videos. The proposed method produces the frames with more realistic textures. (\textbf{Zoom-in for best view})
		}
		\label{fig:supp-visual-compare}
	\end{figure*}

	In the main paper, we have demonstrated the effectiveness of the proposed SATB strategy and the state-of-the-art performance of SATB-VR. Here, we provide expanded qualitative results to further validate our claims.
	Specifically, Fig.~\ref{fig:supp-visual-artifact} presents additional examples illustrating the indispensability of the SATB strategy. By explicitly resolving the train-inference discrepancy, it effectively eliminates the severe visual artifacts typically caused by naive joint training.
	Furthermore, Fig.~\ref{fig:supp-visual-compare} offers more comprehensive visual comparisons against existing methods across both synthetic and real-world scenarios, where our method recovers more realistic fine textures.
	
	Notably, since temporal consistency and dynamic perceptual quality are best evaluated through continuous frames, we highly encourage readers to view the side-by-side video comparisons available on our project page: \url{https://github.com/chenxx89/SATB-VR}.

\end{document}